\documentclass[aip,jmp,amsmath,graphics,preprint,author-year]{revtex4-1}

\newcommand{\be}{\begin{equation}}
\newcommand{\ee}{\end{equation}}
\newcommand{\bc}{\begin{center}}
\newcommand{\ec}{\end{center}}
\newcommand{\bd}{\begin{description}}
\newcommand{\ed}{\end{description}}
\newcommand{\bi}{\begin{itemize}}
\newcommand{\ei}{\end{itemize}}
\newcommand{\bz}{\bs{z}}
\newcommand{\bx}{\bs{x}}
\newcommand{\bk}{\bs{k}}

\newcommand{\bxx}{\bs{X}}

\newcommand{\bt}{\bs{\theta}}

\newcommand{\bp}{\bs{\phi}}
\newcommand{\bs}{\boldsymbol}

\usepackage{graphicx}
\usepackage{amsfonts}
\usepackage{appendix}
\draft 

\usepackage{xcolor}

\begin{document}


\title[Interpretable reduced-order modeling with time-scale separation]{Interpretable reduced-order modeling with time-scale separation} 



\author{Sebastian Kaltenbach}
   \email{skaltenbach@seas.harvard.edu}
\affiliation{ 
ETH Zurich CSE-Lab and Harvard SEAS
}%
\author{Phaedon-Stelios Koutsourelakis}%
\affiliation{ 
Professorship of Data-driven Materials Modeling, TUM
}%

\author{Petros Koumoutsakos}
\affiliation{%
Harvard SEAS
}%


\date{\today}

\begin{abstract}
Partial Differential Equations (PDEs) with high dimensionality
are commonly encountered in computational physics and engineering. However, finding solutions for these PDEs can be computationally expensive, making model-order reduction crucial. We propose such a data-driven scheme that automates the identification of the time-scales involved and,
can produce stable predictions forward in time as well as under different initial conditions not included in the training data. 
To this end, we combine a non-linear autoencoder architecture with a time-continuous model for the   latent dynamics in the complex space. 
 It readily allows for the inclusion of sparse and irregularly sampled training data. The learned,  latent dynamics are interpretable and reveal the different temporal scales involved. We show that this data-driven scheme can automatically learn the independent processes that decompose  a system of linear ODEs along the eigenvectors of the system's matrix. 
 Apart from this, we demonstrate the applicability of the proposed framework in a hidden Markov Model and the (discretized) Kuramoto-Shivashinsky (KS) equation. Additionally, we propose a probabilistic version, which captures predictive uncertainties and further improves upon the results of the deterministic framework.
\end{abstract}


\maketitle 

\section{Introduction}
High-fidelity simulations of critical phenomena such as ocean dynamics and epidemics have become essential for decision-making.
They  are based on physically-motivated PDEs expressing system dynamics that span multiple spatiotemporal scales and which necessitate cumbersome  computations \citep{givon_extracting_2004}. 
In recent years there is increased attention to the development of data-driven models that can accelerate the solution of these PDEs \citep{alber2019integrating} as well as reveal salient, lower-dimensional features that control the long-term evolution.
In most cases, data-driven reduced-order models are  not interpretable. In particular, models based on neural networks  despite good predictive capabilities\citep{chen2018neural,li2020scalable,vlachas2022multiscale}, they offer a black-box description of the system dynamics. A possible remedy is applying a symbolic regression to the learned neural network representation \citep{cranmer2020discovering}, but this adds additional computational cost due to the two-step procedure. A number of frameworks such as SINDy \citep{brunton2016discovering} allows to learn interpretable dynamics but it relies on the a-priori availability of lower-dimensional descriptors and of time-derivatives which can be very noisy for both simulation and experimental data. Other frameworks are tailored to specific problems such as molecular dynamics \citep{noe_machine_2020}. Here, we present  a framework that only needs the value of the observables, and not their derivatives, as training data and is capable of  identifying interpretable latent dynamics. The deployment of interpretable latent dynamics  ensures that conservation of  important properties of that are reflected in the reduced-order model \citep{kaltenbach2020incorporating,kaltenbach2021physics}. The present method  is related to approaches based on the Koopman-operator \citep{koopman_hamiltonian_1931,budivsic2012applied,klus_data-driven_2018} extended Dynamic Mode Decomposition (eDMD) \citep{williams2015data} but uses continuous complex-valued latent space dynamics and only requires one scalar variable per latent dimension to describe the latent space dynamics. Therefore we do not have to enforce any parametrizations on the Koopman matrix \citep{pan2020physics}. The time-continuous formulation moreover allows to incorporate sparse and irregularly sampled training data and fast generation of predictions after the training phase. By using a complex-valued latent space we can also incorporate harmonic effects  
and reduce the number of latent variables needed.\\
Linear and non-linear  autoencoders are used to  map the observed, high-dimensional time-series to the lower-dimensional, latent representation \citep{kingma2013auto,champion_data-driven_2019} and we identify simultaneously the autoencoder as well as the latent dynamics by optimizing a combined loss function. Hence the to tasks  of dimensionality reduction and discovery of the reduced dynamics are unified 
\citep{horenko_automated_2008,horenko_simultaneous_2008} while other frameworks treat the two parts separately \citep{vlachas2022multiscale}. Apart from  using an architecture based on autoencoders to identify the latent space, projection-based methods could also be employed \citep{snyder_reduced_2022}.  \\
We are also proposing a probabilistic version of our algorithm \citep{kaltenbach2021physics} that makes use of probabilistic Slow Feature Analysis \citep{turner_maximum-likelihood_2007,zafeiriou_probabilistic_2015}. This allows for a latent representation that arart from being time-continuous, can  quantify the predictive uncertainty and hierarchically decompose the dynamics 
 into their pertinent scales while promoting the discovery of slow processes that control the system's evolution over long time horizons. 

The rest of the paper is structured as follows: We introduce the methodological framework as well as algorithmic details in section \ref{sec:methodology}. Particular focus is paid on the interpretability of the inferred lower-dimensional dynamics. In section \ref{sec:examples} we present  three numerical illustrations, i.e. a system of linear ODEs, a hidden Markov Model and the discretized KS-equation. We then present in section \ref{sec:prob} the  probabilistic extension of the framework and apply it to the KS-equation. We conclude with a summary and a short discussion about possible next steps.

\section{Methodology}
\label{sec:methodology}
We introduce the autoencoders deployed in this  work, followed by the interpretable latent space dynamic and discuss the training process.\\
We consider data from  high-dimensional time series $\boldsymbol{x}_n \in \mathbb{R}^f$ with $n=1,...,T$. We remark that the intervals between the different states do not need to be uniformly spaced.\\
\subsection{Autoencoder}
A core assumption of the method is that each high-dimensional state $\boldsymbol{x}_n$ can be compressed to a lower-dimensional representation $\boldsymbol{z}_n \in \mathbb{C}^c$ with $c<<f$. We identify this  lower-dimensional representation by an  autoencoder consisiting of a parameterized encoder and decoder. The encoder maps the high-dimensional representation to the latent space as: 
\begin{equation}
    \boldsymbol{z}_n=\boldsymbol{E}_{\boldsymbol{\theta}}(\boldsymbol{x}_n)
\end{equation}
The latent space is complex-valued. 
The decoder reconstructs the high-dimensional representation based on the latent variables as: 
\begin{equation}
    \boldsymbol{x}_n=\boldsymbol{D}_{\boldsymbol{\theta}}(\boldsymbol{z}_n)
\end{equation}
We denote the parameters of the encoder as well as the decoder by $\boldsymbol{\theta}$. As discussed later in Section \ref{sec:train}, both set of parameters are optimized simultaneously during training and therefore there is no need for differentiating them.

\subsection{Interpretable Latent Space Dynamics}
We employ a propagator in the latent space to capture the reduced-order dynamics of the system. In contrast to other time-extended variational autoencoder frameworks, our representation uses complex valued latent variables. In addition the latent variables are treated independently. The latter feature enables us to have an interpretable latent dynamics as well as a model that is especially suitable for being trained in the Small Data regime due to the small number of required parameters. This is in contrast to temporal propagators such as  LSTMs \citep{hochreiter1997long}.\\

For each dimension $i$ of the latent variable z we are using the following continuous ODE in the complex plane:

\begin{equation}
\frac{d z_i}{d t} = - \lambda_i  z_i
\end{equation}
By solving this ODE, we can define the operator:
\begin{equation}
\boldsymbol{z}_{n+\Delta t_n}=\boldsymbol{P}_{\lambda}(\boldsymbol{z}_{n},\Delta t_n) = \boldsymbol{z}_n \odot \exp(-\boldsymbol{\lambda} \Delta t_n)
\label{eq:operator}
\end{equation}
Here, $\boldsymbol{\lambda}$ is a vector containing all the individual $\lambda$'s and $\Delta t_n$ indicates the time-step between the latent states. The symbol $\odot$ is used to indicate a component-wise multiplication. We remark that the latent variables and the parameter governing the temporal evolution are complex numbers and their role in describing the system  dynamics is similar to that of an eigenvalue. The real part is associated with growth and decay whereas the imaginary part is representing the periodic component.
This approach has similarities with the Koopman-operator based methods \citep{koopman_hamiltonian_1931,mezic2013analysis,rowley2009spectral} and  the extended dynamic mode decomposition \citep{williams2015data}. In contrast to the methods mentioned before we are using a continuous formulation in the latent space that allows us to incorporate scarce and irregularly sampled training data and directly rely on complex numbers in the latent space.\\

\subsection{Training and Predictions}
\label{sec:train}
We optimize a loss function that combines both a reconstruction loss as well as a loss associated with the error of our learned propagator in the latent space:
\begin{equation}
L= \frac{1}{T} \sum_{n=1}^T \left|\boldsymbol{x}_n -\boldsymbol{D}_{\bt}(\boldsymbol{E}_{\bt}(\boldsymbol{x}_n))\right|^2 + \frac{1}{T-1} \sum_{n=1}^{T} \left| \boldsymbol{E}_{\bt}(\boldsymbol{x}_{n+1})-\boldsymbol{P}_{\lambda}(\boldsymbol{E}_{\bt}(\boldsymbol{x}_n),\Delta t_n) \right|^2
\end{equation}

We note that we could directly incorporate mini-batch training by only taking the summation over a subset of the $N$ available training data. 
For new predictions of unseen states, we use the encoder to generate a latent representations which is then advanced in time by the learned propagator. At a designated time step we are using the decoder to reconstruct the high-dimensional solution.

\section{Numerical Illustrations}
\label{sec:examples}
We applied our algorithm to three systems. First, we show that the algorithm is capable of exactly reproducing the solution of a linear ODE and to identify its eigenvalues. Afterwards we are applying the framework to a high-dimensional process generated by a complex latent dynamics, which is correctly identified. As a final test case, we are applying the algorithm to a Kuramoto Shivashinski (KS) equation. 

\subsection{Linear ODE}
\begin{figure}
    \centering
    \includegraphics[scale=0.5]{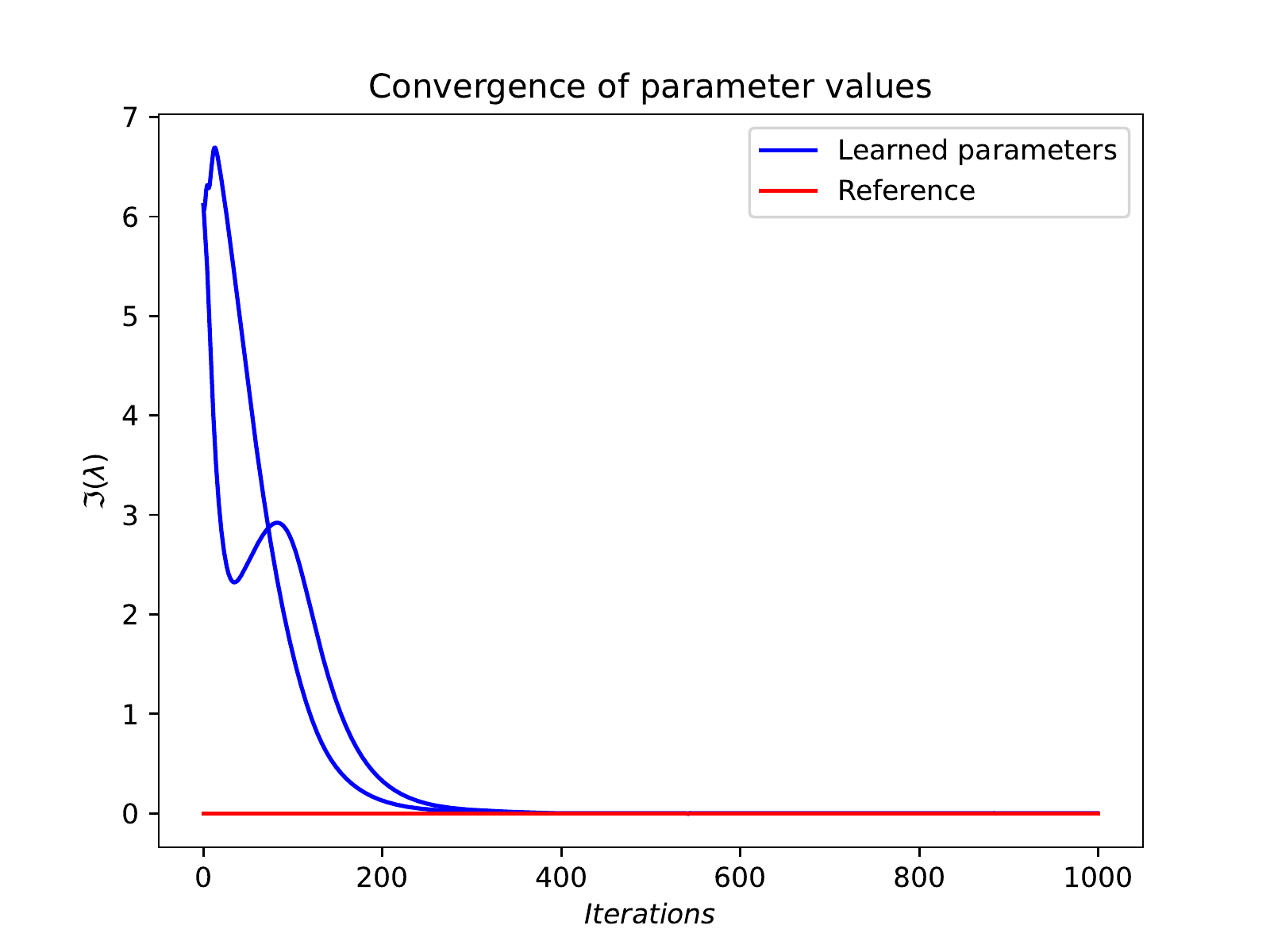}
    \includegraphics[scale=0.5]{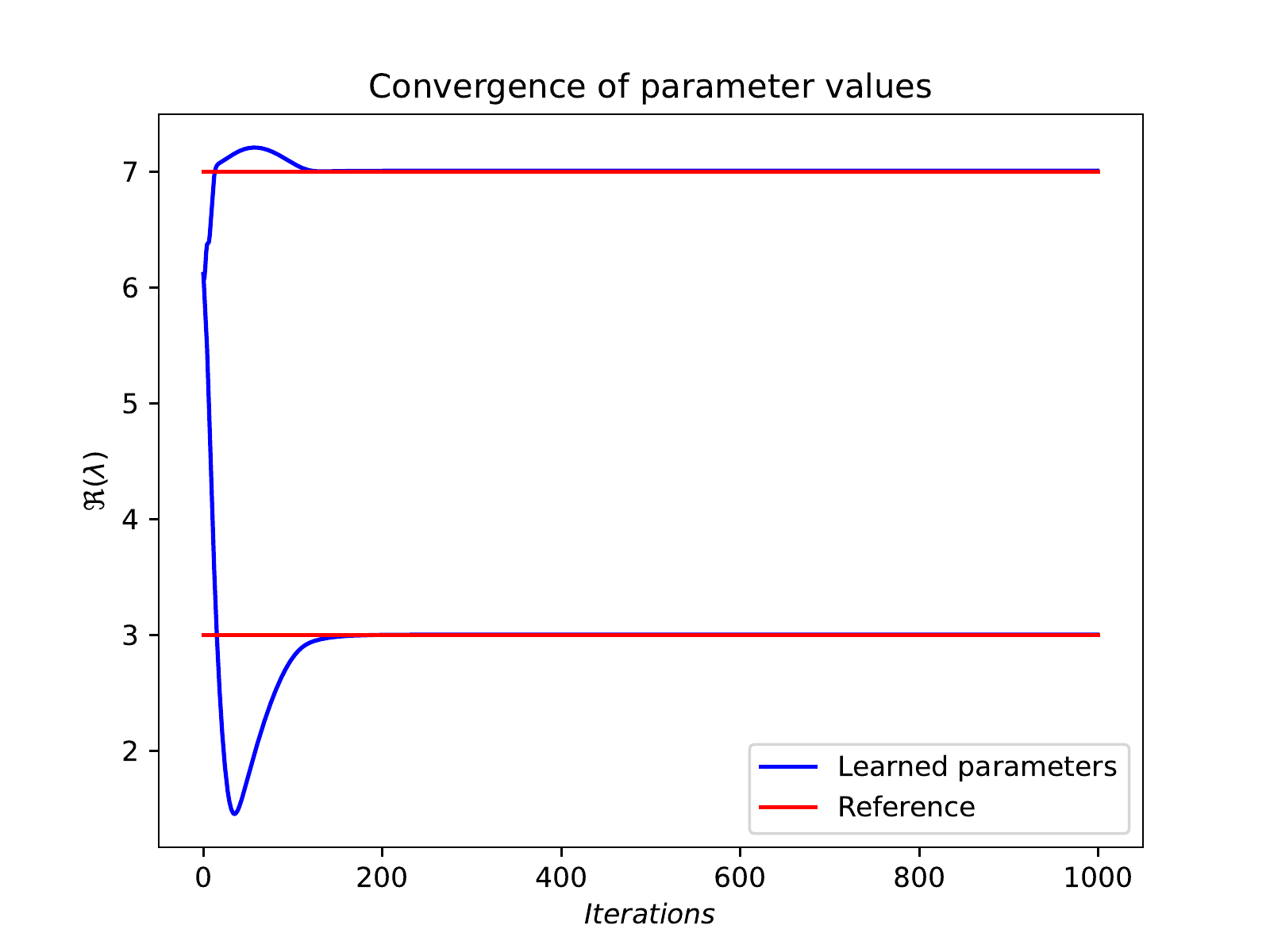}
    \caption{Convergence of $\bs{\lambda}$ for the linear ODE test case}
    \label{fig:ode}
\end{figure}

We are considering a two-dimensional ODE system for $\bx= \begin{pmatrix} y_1 & y_2 \end{pmatrix}$:
\begin{equation}
    \begin{pmatrix} \dot{y_1} \\ \dot{y}_2 \end{pmatrix}= \begin{pmatrix} -5 & 2 \\ 2 &-5 \end{pmatrix} \begin{pmatrix} y_1 \\ y_2 \end{pmatrix}
\end{equation}
with initial conditions $y_1=10$ and $y_2(0)=-3$. Training data is generated by solving the ODE using a Runge-Kutta-solver for $t_{max}$=2  with a time-step size of $\Delta t=2.5\cdot 10^{-4}$  and then sampling 40 time-series with 150 data points each from the obtained trajectory. Hence the training data consists of:
\bi
\item $40$ time-series
\item with each consisting $150$ observations of the $\bx$ with varying time-steps $\Delta t$
\ei
Based on the obtained training data we run our algorithm using a linear encoder and decoder structure as well as two latent variables $\boldsymbol{z}$. The loss function was optimized using the Adam algorithm \citep{kingma2014adam}.\\
As we consider  a linear ODE  we can analytically compute the eigenvalues involved and compare it with the parameters $\lambda$ identified by our algorithm. We observe in Figure \ref{fig:ode} that the algorithm was able to recover the correct values, i.e. the eigenvalues $7$ and $3$ of the given linear ODE. The system does not have a periodic component and  the two imaginary parts correctly go to zero, whereas the real parts converge to the reference value.

Moreover we are also able to identify for the linear mapping between our latent variables $z$ and the training data a matrix consisting of a multiple of the eigenvectors (1,1) and (1,-1) and thus the correct solution. This example was chosen to show that the algorithm is able to quickly identify the exact solution of a linear ODE in terms of its linearly independent components.

\subsection{Hidden multiscale dynamics}

We consider eight-dimensional synthetic time series data produced by an underlying two-dimensional complex valued process. In particular, the data points $\bx$ are generated by first solving for the temporal evolution for the two complex-valued processes $p_1$ and $p_2$ and than mapping to the eight-dimensional space by using a randomly sampled linear mapping $\boldsymbol{W}$. One of the two processes used to generate the data is chosen to be much slower than the other one and both processes have a periodic component.
\begin{align}
      \frac{dp_1}{dt} &= (-0.1+1i)p_1 \\
      \frac{dp_2}{dt} &= (-0.9+1.5i)p_2 \\
      \boldsymbol{x} &=\boldsymbol{W}\begin{pmatrix} p_1 \\p_2 \end{pmatrix}
\end{align}

As training data we consider  40 time series with 150 data points each, obtained by simulating the described processes for a maximum of $t=15$ s and then sampling from the obtained data points.
Hence the training data consists of:
\bi
\item $40$ time-series
\item with each consisting $150$ observations of the $\bx$ at a uniform time-step $\Delta t=0.0025$
\ei

\begin{figure}[h!]
    \centering
    \includegraphics[scale=0.5]{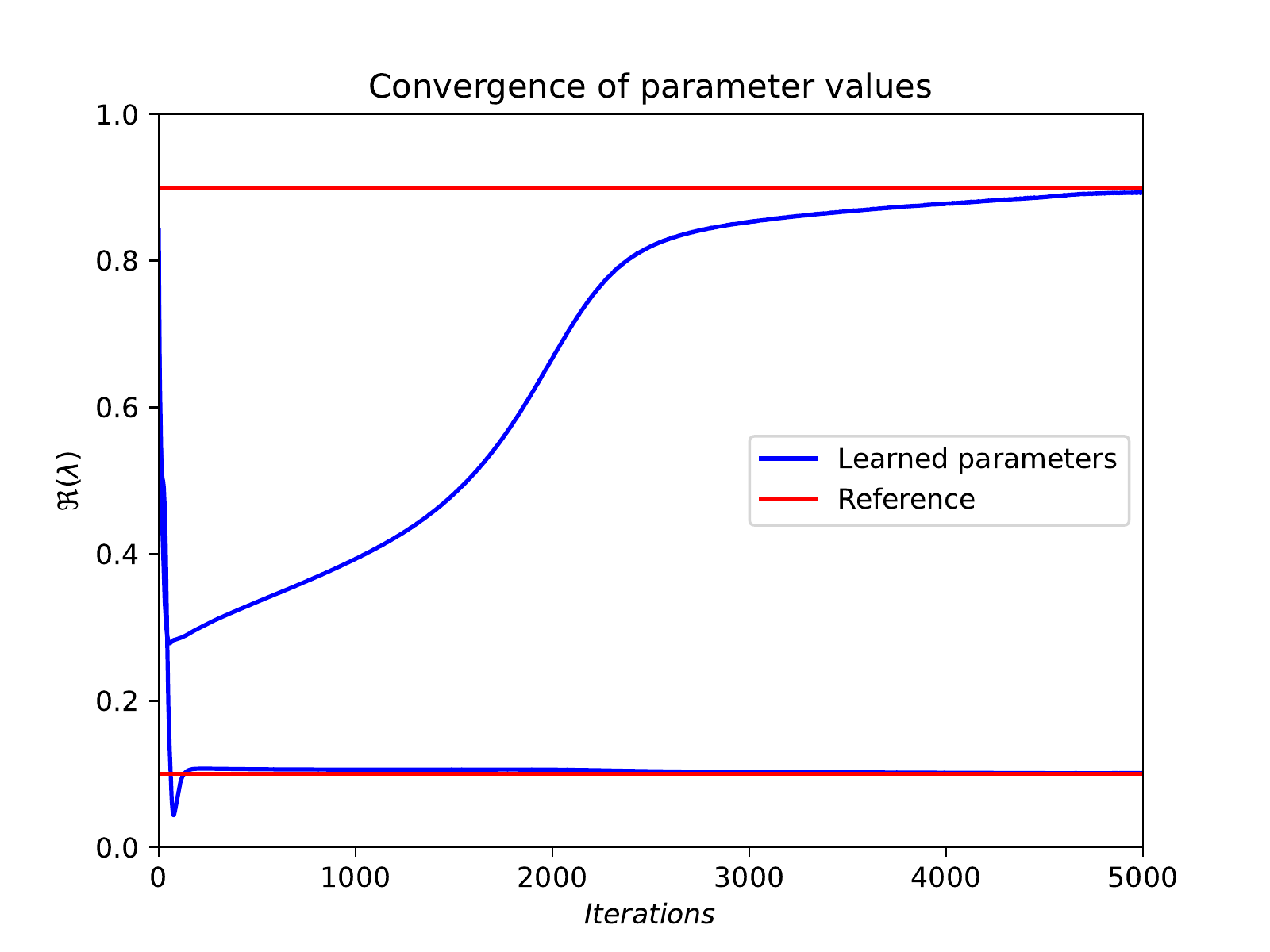}
    \includegraphics[scale=0.5]{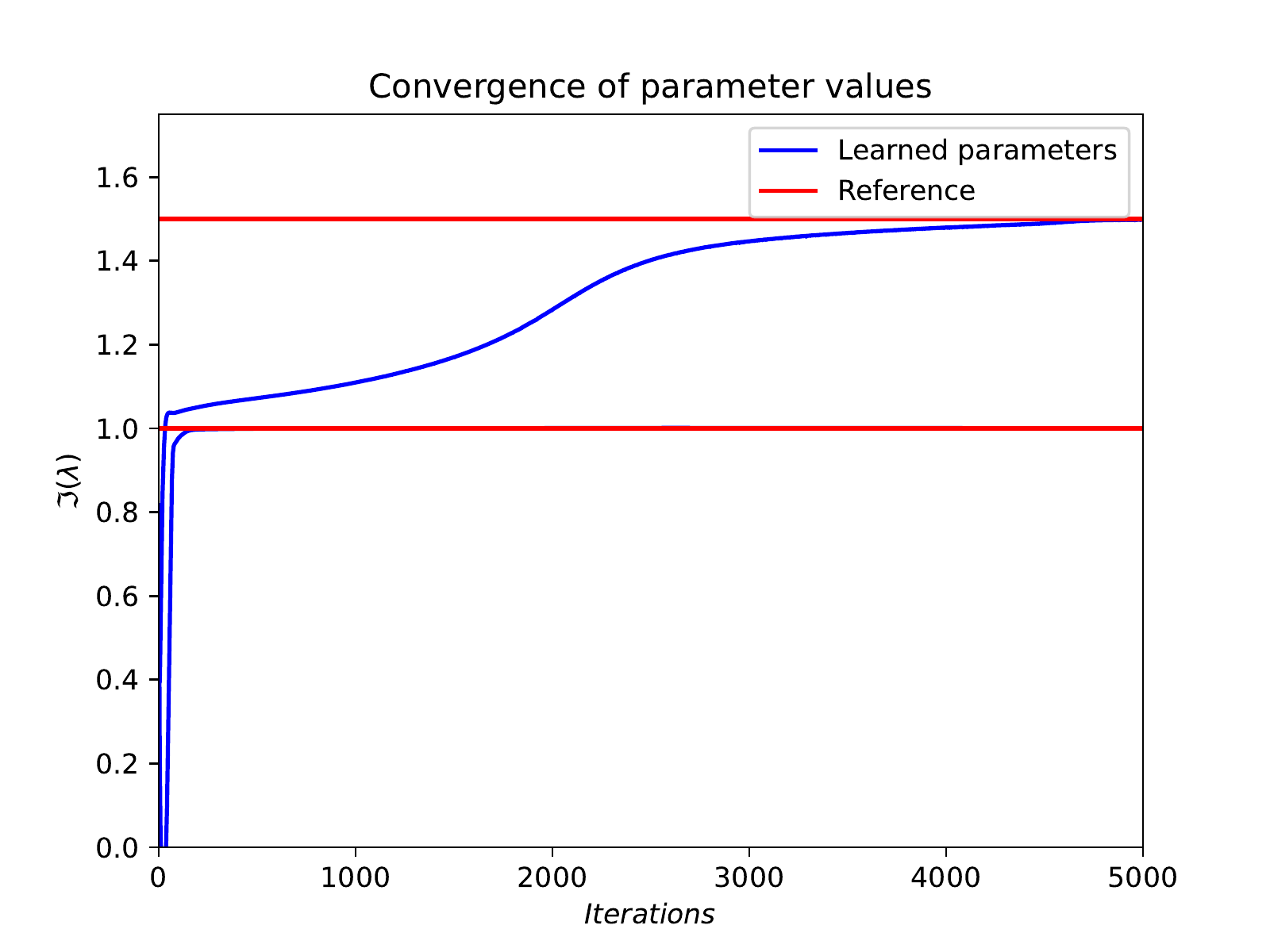}
    \caption{Convergence of the obtained real parts for the parameters for the latent dynamics. Both are converging to the reference value. We note that the slower process is converging significantly faster than the faster process.}
    \label{fig:lambda_HMM}
\end{figure}

The autoencoder obtained consists of one linear layer for both the decoder as well as the encoder.\\
The model is trained for 5000 iterations using the Adam optimizer and a learning rate of $10^{-3}$. The results for the convergence of the parameters $\lambda_1$ and $\lambda_2$ can be found in Figure \ref{fig:lambda_HMM}. We note that the process which is slower decaying and thus more responsible for the long-term evolution of the system has a higher convergence rate than the faster process.

\begin{figure}[p]
    \centering
    \includegraphics[scale=0.4]{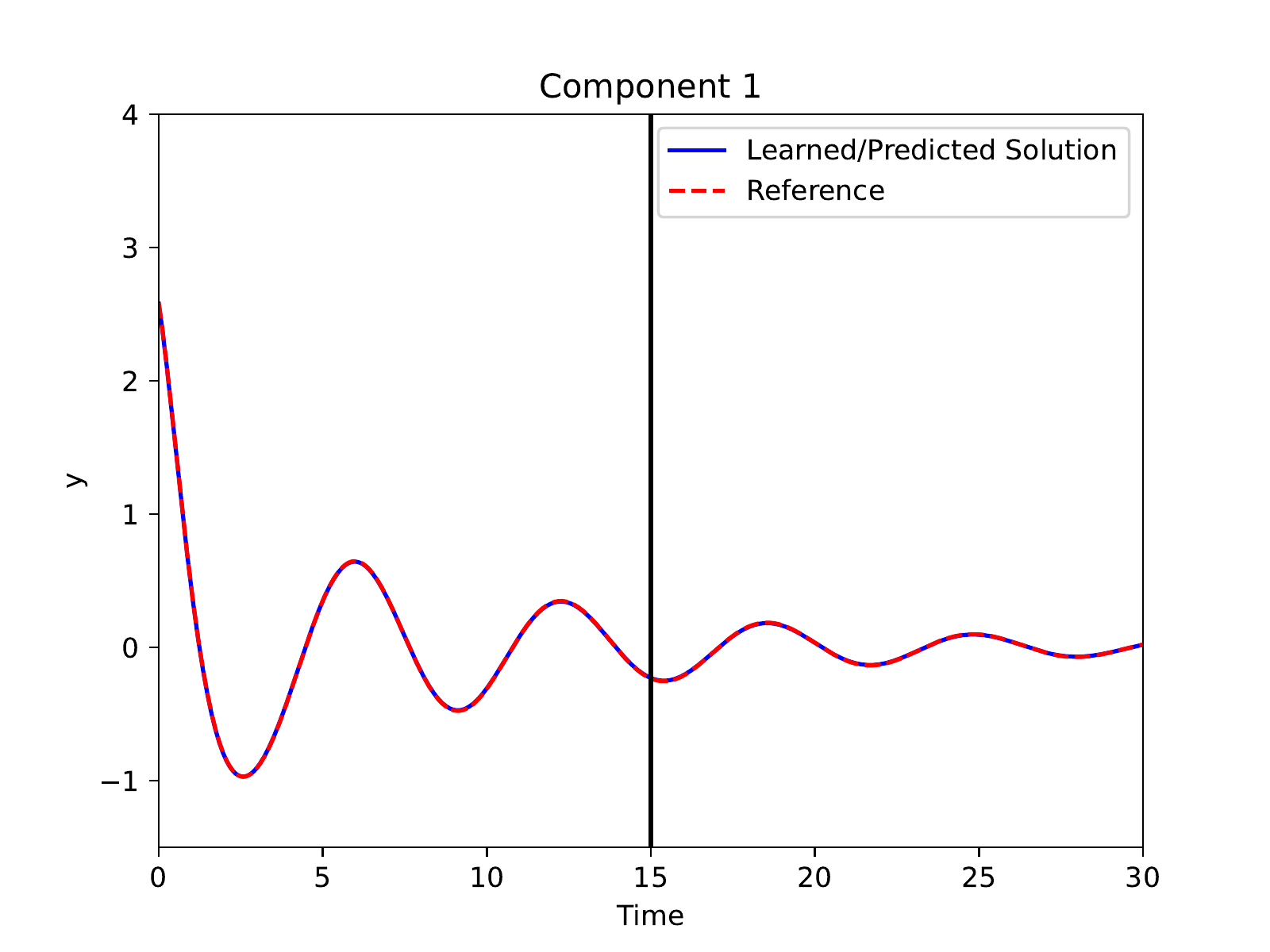}
    \includegraphics[scale=0.4]{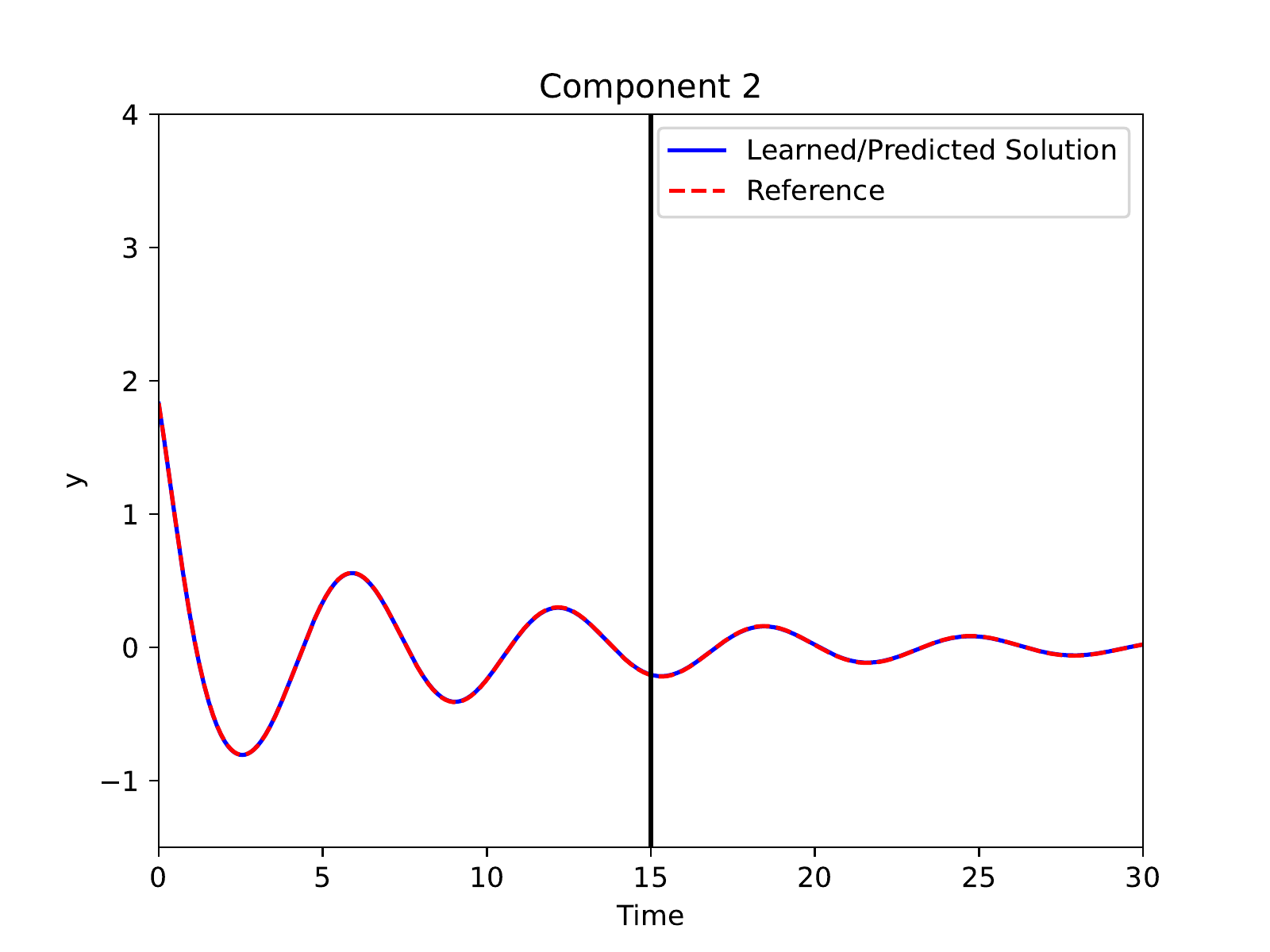}\\
    \includegraphics[scale=0.4]{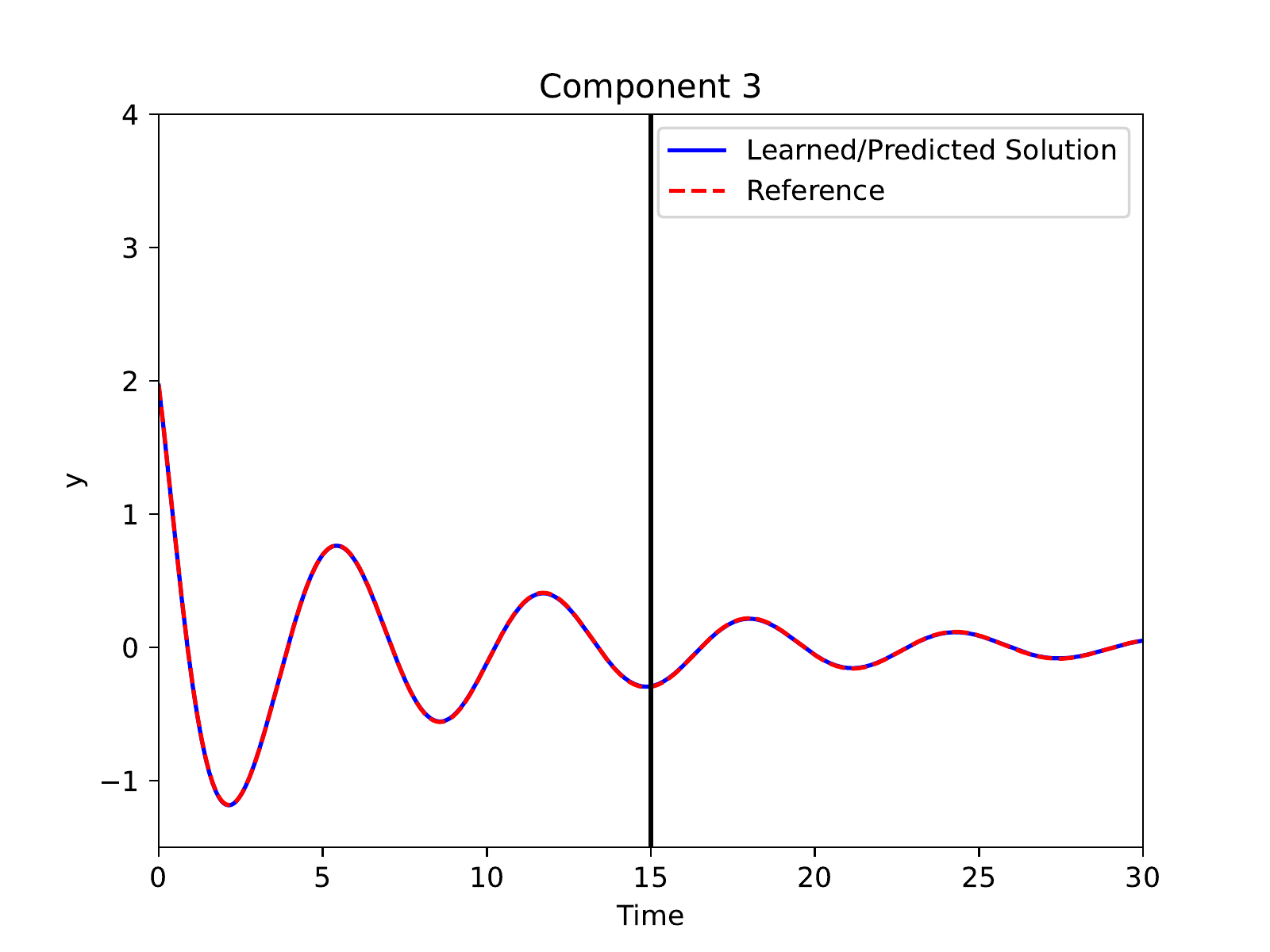}
    \includegraphics[scale=0.4]{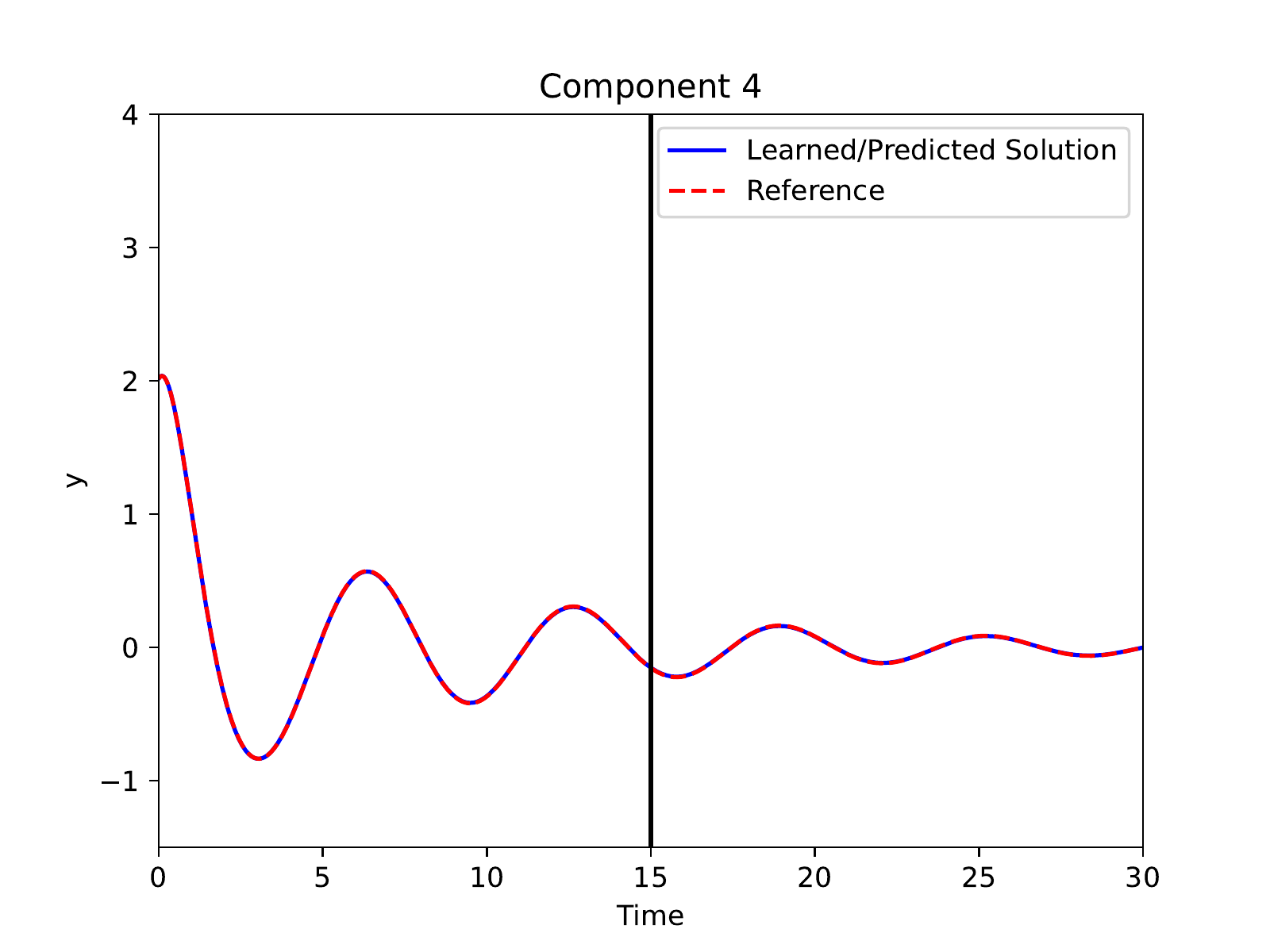}\\
    \includegraphics[scale=0.4]{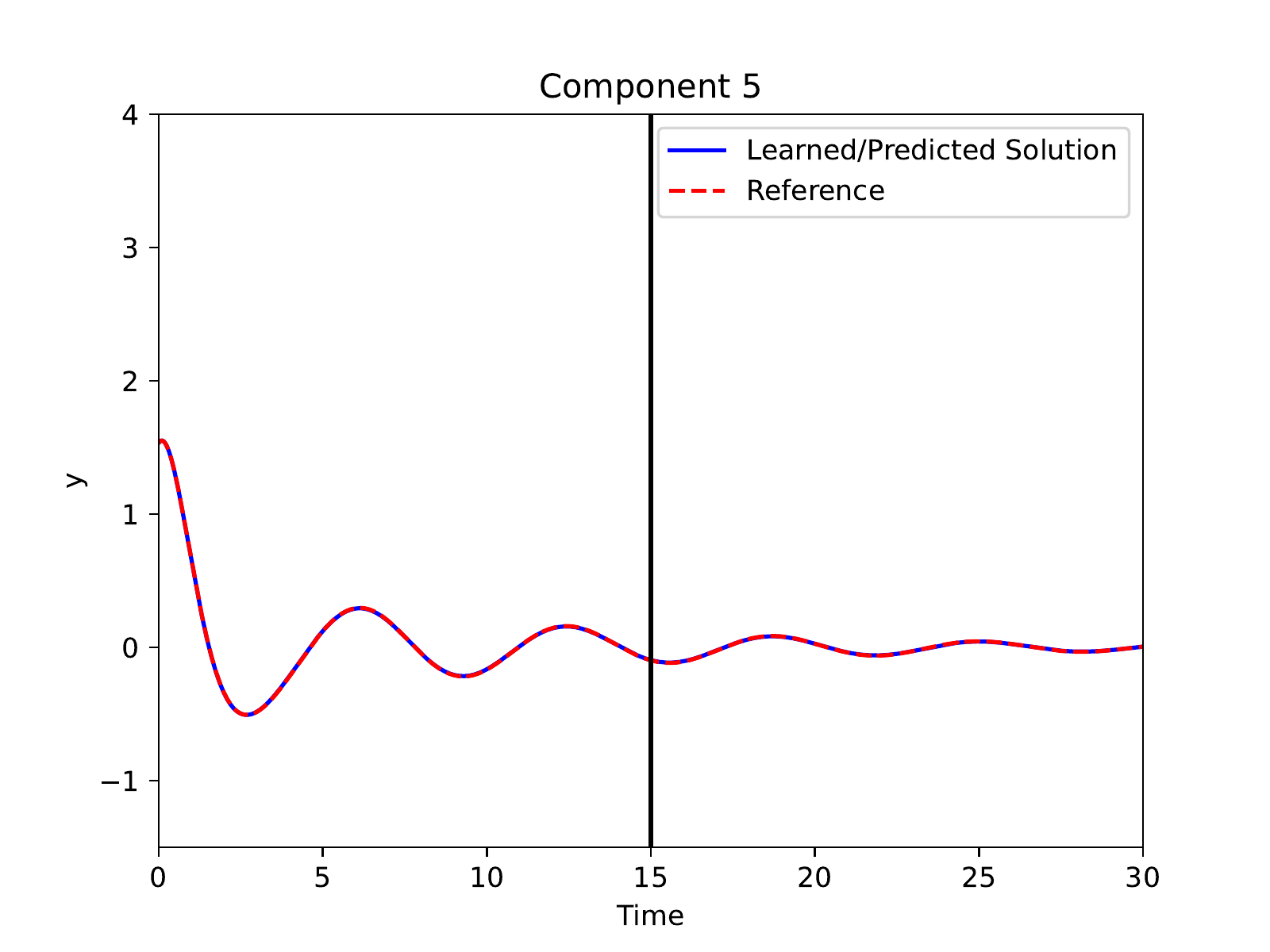}
    \includegraphics[scale=0.4]{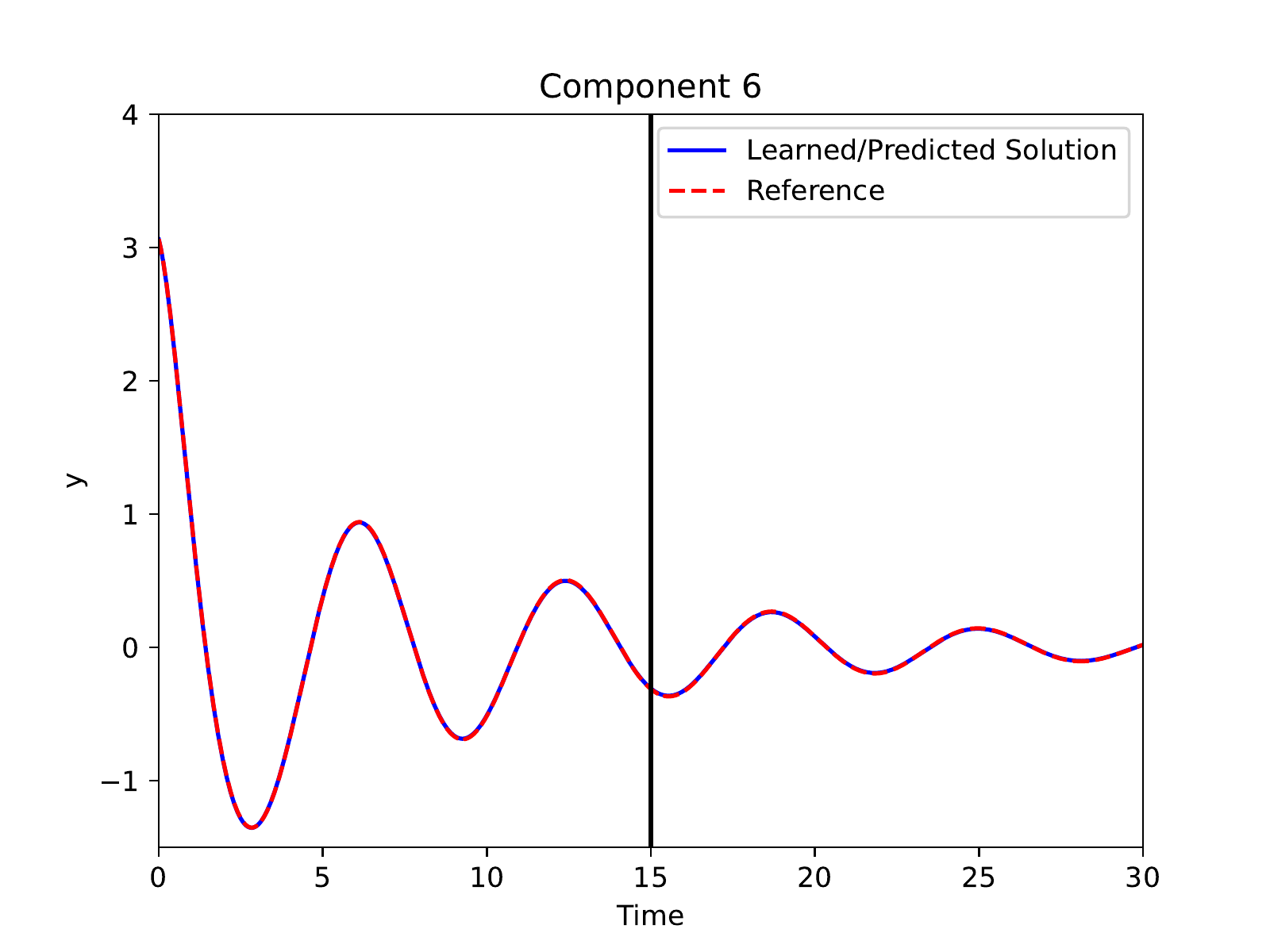}\\
    \includegraphics[scale=0.4]{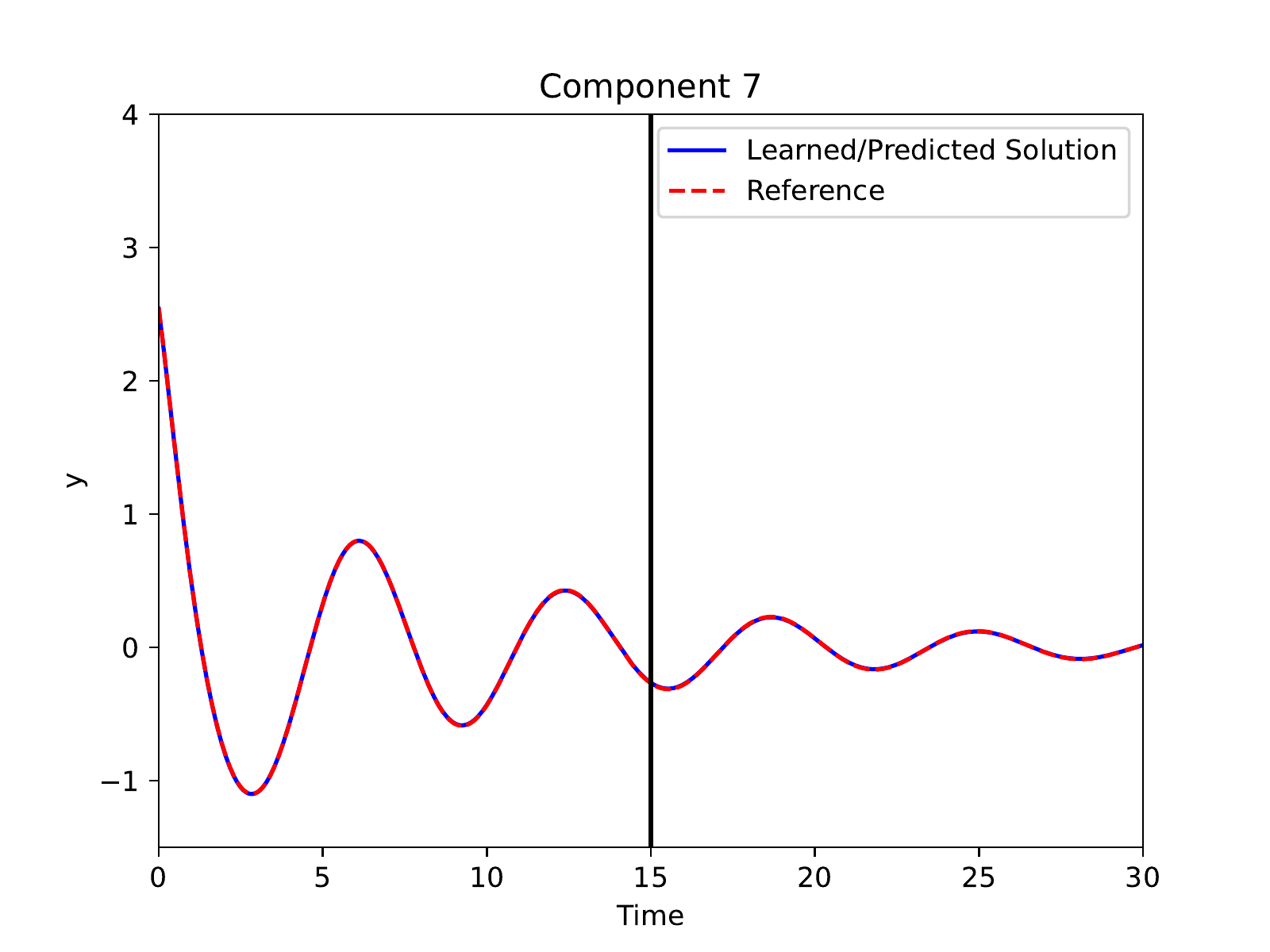}
    \includegraphics[scale=0.4]{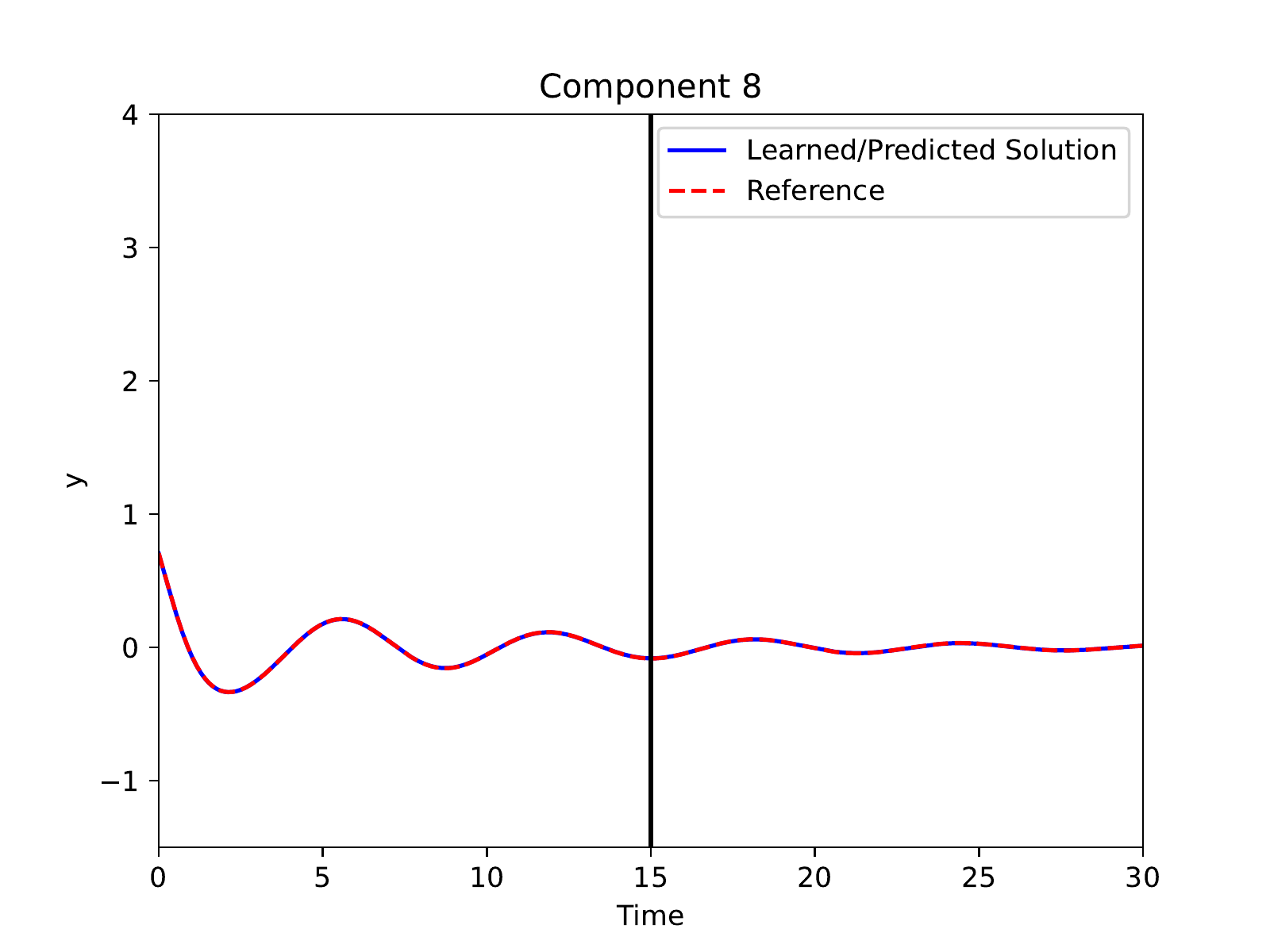}\\
    \caption{Reconstructed and predicted evolution for all components of the eight-dimensional system. The black lines divides the area for which training data was given from the area without raining data.}
    \label{fig:HMM}
\end{figure}

With the obtained parameters $\lambda$ as well as the trained autoencoder, we compute predictions based on the last time step used for training, i.e. we apply the encoder to obtain our latent representation and than use the latent dynamics to advance the latent representation in time. Afterwards, we employ the decoder to reconstruct the full high-dimensional system. The results can be found in Figure \ref{fig:HMM} and show very good agreement between predictions and reference data.

This example shows that our model is successfully able to carry out dimensionality reduction and moreover indicates that the convergence rate between latent processes can be different. The latter is relevant when training models as for accurate predictions all latent processes and their dynamics should be converged.

\subsection{Kuramoto-Sivashinsky}
Finally, we applied our algorithm to the KS equation and aim to identify a reduced-order model for the solution $u(y,t)$:
\begin{equation}
    \frac{\partial u}{\partial t} = -\mu \frac{\partial^4 u}{\partial y^4}- \frac{\partial^2 u}{\partial y^2}- u\frac{\partial u}{\partial y}
\end{equation}
We employed periodic boundary conditions, $\mu=1$ and a domain size $y \in [0,22]$. For this domain-size, the KS-equation exhibits a structurally stable chaotic attractor as discussed in \cite{robinson1994inertial,kassam2005fourth}. 
The equation is discretized in space using a discretization step of $\frac{22}{64}$ resulting in a state vector $\bx$ of dimension 64 and a nonlinear system of coupled ODEs. 
This is solved using a stiff fourth-order solver \citep{kassam2005fourth} and 1000 data-points within the first 250 seconds of the solution are divided into 40 time-series with 25 points each to be used as training data. Hence the training data consists of:
\bi
\item $40$ time-series
\item with each consisting $25$ observations of  $\bx$ at a uniform time-step $\Delta t=0.25$
\ei

\begin{figure}[h]
    \centering
    \includegraphics[scale=0.4]{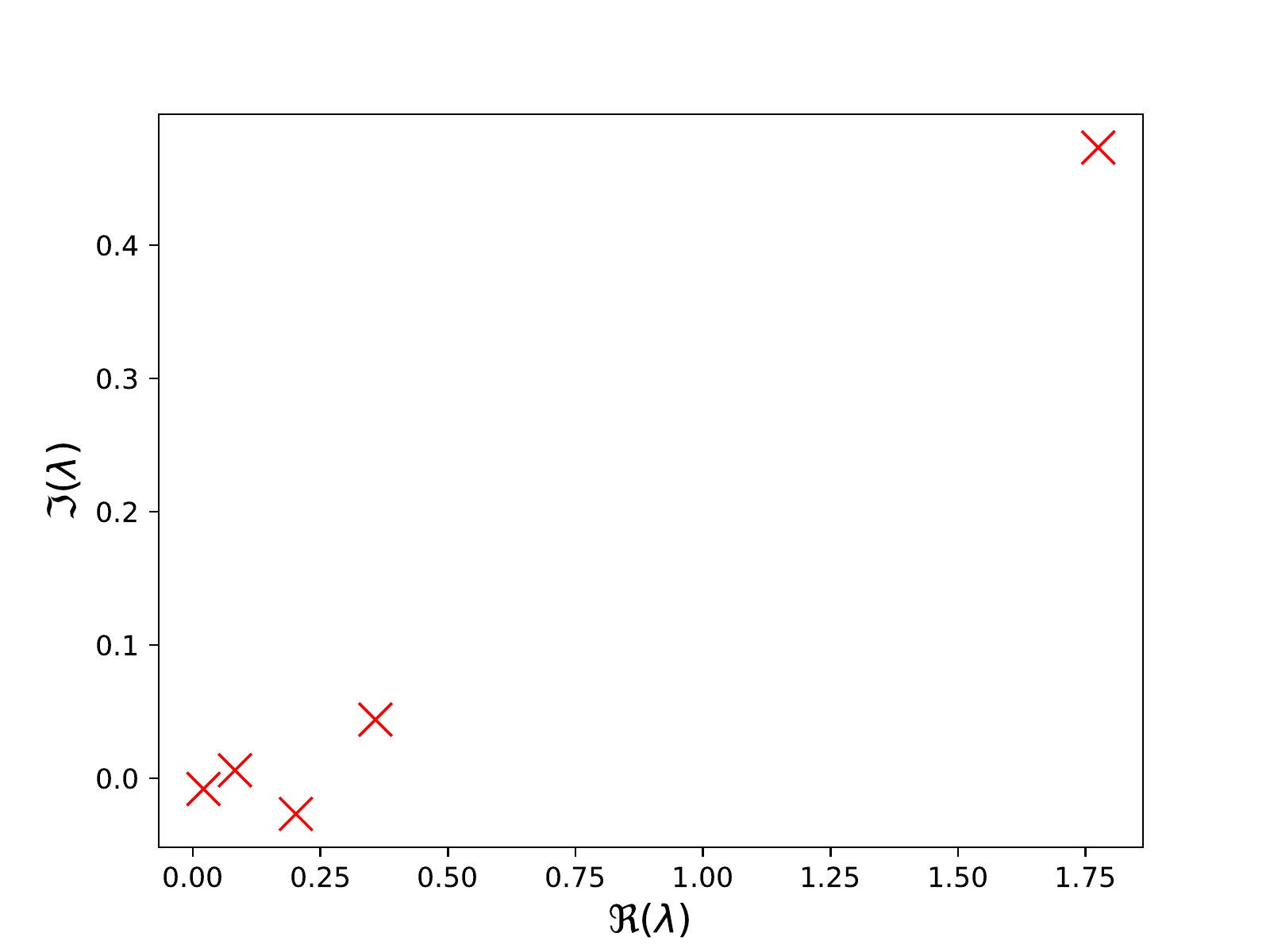}
    \caption{Obtained parameters $\lambda$ for the KS-equation}
    \label{fig:KS_Lambda}
\end{figure}

We employed a non-linear encoder and decoder with four fully-connected layers each and ReLU-activation functions as well as Dropout Layers \citep{srivastava2014dropout} between the fully-connected layers. We trained  the model for 200000 iterations using Adam and a learning rate of $5\cdot 10^4$ and assuming a five-dimensional latent space.  
We obtained the $\lambda$'s in Figure \ref{fig:KS_Lambda}. Four latent variables have $\lambda$'s close to zero and thus a slow temporal dynamic that is responsible for the long-term evolution whereas one latent variable is quickly decaying.

\begin{figure}[h]
    \centering
    \includegraphics[scale=0.4]{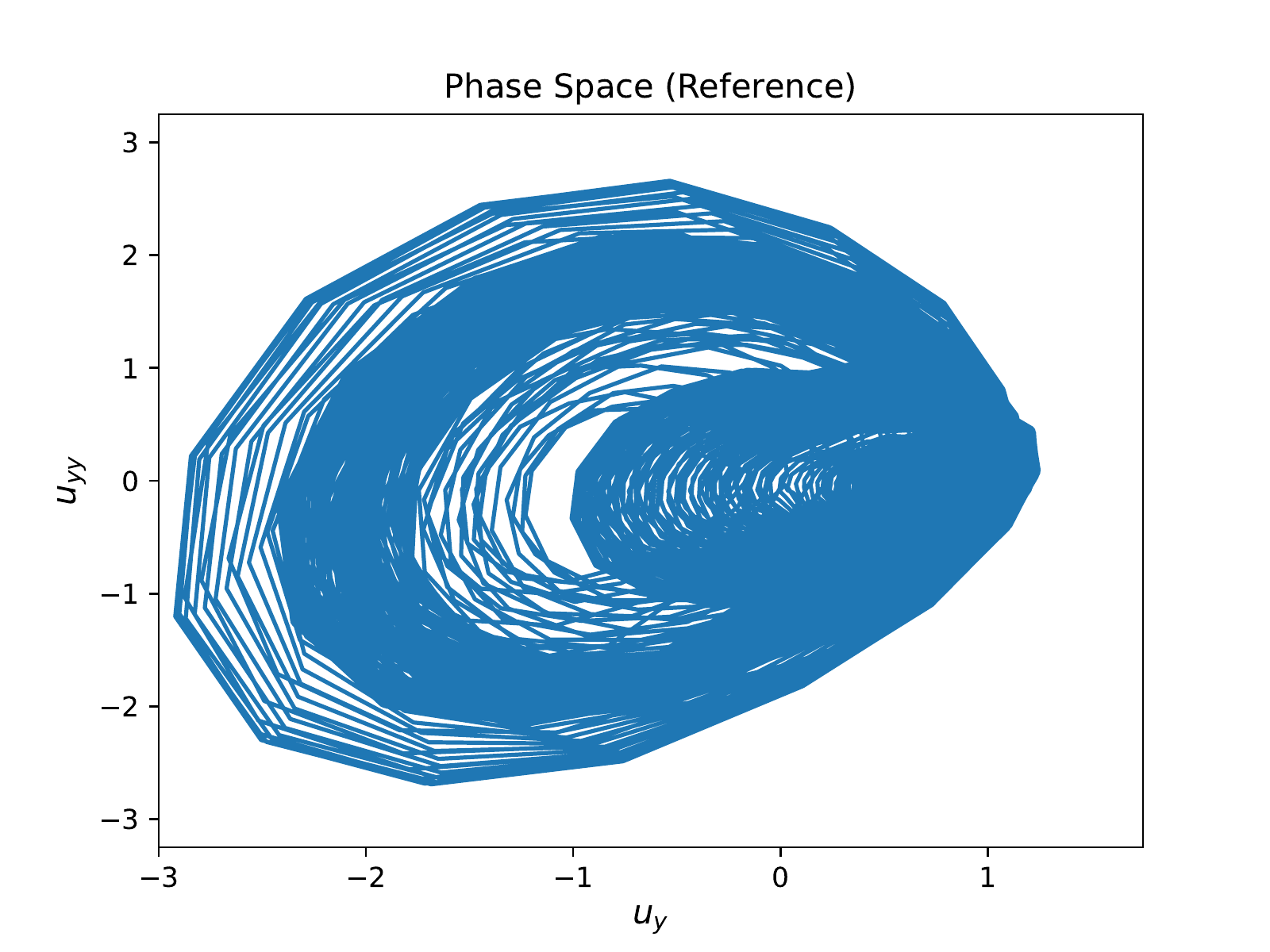}
    \includegraphics[scale=0.4]{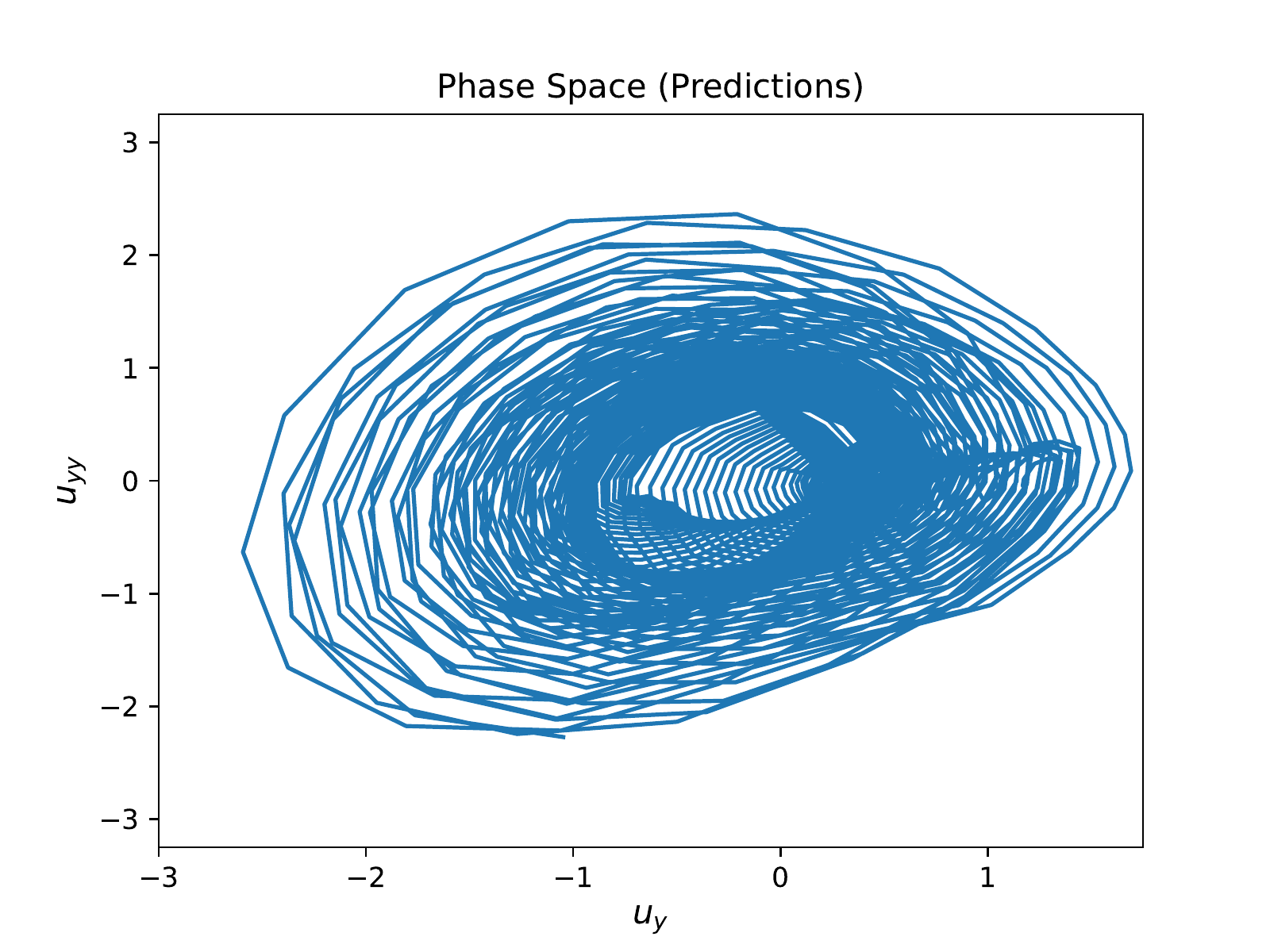}
    \caption{Comparison between the phase-space of the reference solution (left) and the phase-space of the predictions}
    \label{fig:KS_det}
\end{figure}

Based on the obtained parameters, we do predictions based on an unseen initial condition not contained in the training data. We are able to reconstruct the correct phase space based on our predictions despite only using a very limited amount of training data.  The results for the phase space can be seen in Figure \ref{fig:KS_det}. Although the small-scale fluctuations in the temporal dynamics are not well captured, the model identifies the correct manifold which has a good accuracy compared to the reference solution. All phase-spaces were obtained by using a finite-difference operator on the data or predictions. These results are in accordance with \cite{vlachas2022multiscale} whose LSTM-based temporal dynamic model was also able to find the correct phase space but not to track the actual dynamics for long-term predictions.
Our model is not able to account for noise in the temporal evolution and thus dealing with chaotic, small-scale fluctuations is challenging. We believe that a probabilistic version of our algorithm could be advantageous here.

\newpage
\section{A probabilistic extension}
\label{sec:prob}

This section contains a fully probabilistic formulation for the deterministic model discussed before. We replace the Autoencoder with a Variational Autoencoder and the ODE in the latent space with a SDE. The loss function which we optimize is the Evidence-Lower Bound (ELBO).
\subsection{Model Structure}
We postulate the following relations for our probabilistic model using an Ornstein-Uhlenbeck (OU) for each dimension $i$ of the latent space and a Wiener process $W_t$ in the latent space:
\begin{align}
   \label{eq:relations}
   d z_{t,i} &= - \lambda_i  z_{t,i}  \: dt + \sigma_i dW_{t,i}   \quad \lambda_i \in \mathbb{C}, ~i=1,2,\ldots, c \\
        \label{eq:2}
    \bx_n &= \bs{G}(\bz_n,\bt)
\end{align}

We again assume that the latent variables $\bz_t$ are complex-valued and a priori independent. Complex variables were chosen as their evolution includes a harmonic components which are observed in many physical systems. 
We assume an initial conditions $z_{0,i} \sim \mathcal{CN}(0,\sigma_{0,i}^2)$. The total parameters associated with the latent space dynamics of our model are thus $\{  \sigma^2_{0,i},\sigma_i^2,\lambda_i  \}_{i=1}^c$ and will be denoted by $\bt$ together with all parameters responsible for the decoder mapping $\boldsymbol{G}$ (see next section).
These parameters along with the state variables $\bz_t$ have to be inferred from the data $\bx_t$.
Based on probabilistic Slow Feature Analysis (SFA) \citep{turner_maximum-likelihood_2007,zafeiriou_probabilistic_2015},   we set  $\sigma_i^2=2;\Re(\lambda_j)$ and $\sigma_{0,i}^2=1$. As  a consequence, a priori, the latent dynamics are stationary. A derivation and reasoning for this choice can be found in Appendix \ref{sec:App_A}. Hence the only independent parameters are the $\lambda_i$, the   imaginary part of which  can account for periodic effects in the latent dynamics.

\subsection{Variational Autoencoder}
We employ a variational autoencoder to account for a probabilistic mappings from the lower-dimensional representation $\bz_n$ to the high-dimensional system $\bx_n$. In particular we are employing a probabilistic decoder
\begin{align}
    \bs{G}(\bz_n,\bt)=p(\bx_n \mid \bz_n, \bt) \quad \text{ and }
\end{align}
The encoder is used to infer the state variables $\bz$ based on the given data and thus defined in the inference and learning section.

\subsection{Inference and Learning}
\label{sec:learning}
Given the probabilistic relations , our goal is to infer the latent variables $\bz_{0:T}$ as well as all model parameters $\bt$. We follow a hybrid Bayesian approach in which the posterior of the state variables is approximated using amortized Variational Inference and  Maximum-A-Posteriori (MAP) point-estimates for $\bt$ are computed.  

The application of  Bayes' rule for each data sequence $\bx_{0:T}$ leads to the following posterior:
\begin{align}
p(\bz_{0:T}, \bt | \bx_{0:T}) = \cfrac{ p(\bx_{0:T} | ,\bz_{0:T}, \bt) ~p(\bz_{0:T}, \bt) }{p(\bx_{0:T}) } \\
= \cfrac{ p(\bxx_{0:T} \mid \bz_{0:T},\bt) ~p(\bz_{0:T} \mid \bt) ~p(\bt) }{p(\bx_{0:T}) }
\label{eq:posterior}
\end{align}
where $p(\bt)$ denotes the prior on the model parameters.
In the context of  variational inference, we use the following factorization of  the approximate posterior
 \begin{align}
     q_{\bp}(\bz_{0:T})= \prod_{n=1}^T  \prod_{i=0}^c q_{\bp}(z_{n,i} \mid \bx_{n})  =\prod_{n=1}^T  \prod_{i=0}^c \mathcal{N}(z_{n,i} | \mu_{n,i}(\bx_{n}),\sigma_{n,i}(\bx_{n}))
 \end{align}
 i.e. we infer only the mean $\mu$ and variance $\sigma$ for each state variable based on the given data points. This conditional density used for inference is the encoder-counterpart to the probabilistic decoder defined in the section before.

It can be readily shown that the optimal parameter values are found by maximizing  the Evidence Lower Bound (ELBO) $\mathcal{F}(q_{\bp}( \bz_{0:T}),\bt)$  which is derived  in Appendix \ref{sec:App_B}. We compute Monte Carlo estimates of the gradient of the ELBO with respect to $\bp$ and $\bt$ with the help of the reparametrization trick \citep{kingma2013auto} and carry out stochastic optimization with  the ADAM algorithm \citep{kingma2014adam}.

\subsection{Results for the probabilistic extension}
\begin{figure}
    \centering
    \includegraphics[scale=0.4]{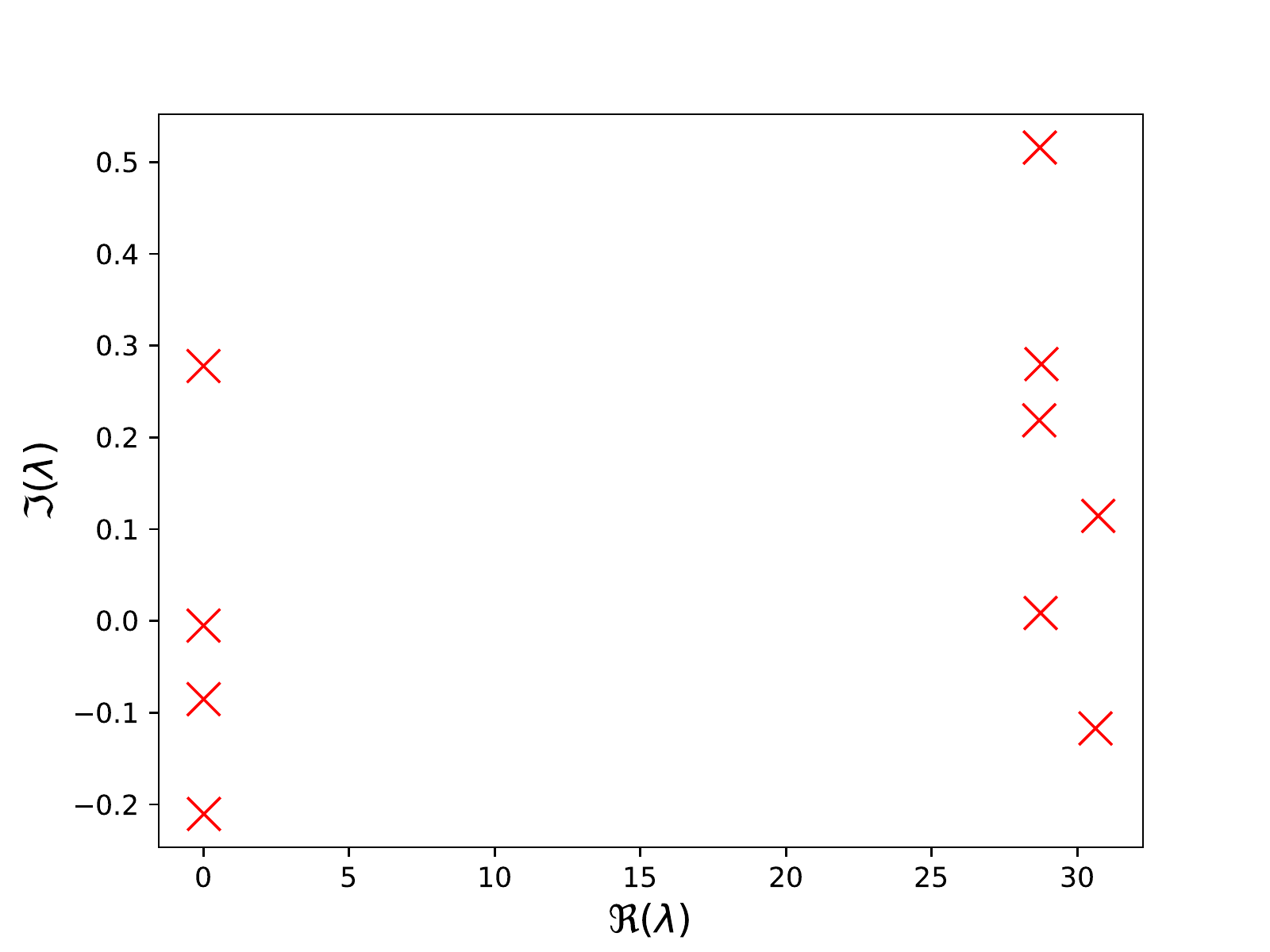}
    \caption{Learned $\lambda$ for the latent dynamics of KS-equation system by using  the probabilistic framework: Similar to the deterministic framework 4 components are slow and responsible for the long-term evolution}
    \label{fig:KS_Lambda_prob}
\end{figure}

We applied our probabilistic version to the KS-equation. We used the same settings as for the deterministic approach but considered up to 10 complex latent variables. The obtained $\lambda$'s are in Figure \ref{fig:KS_Lambda_prob}.

\begin{figure}[p]
    \centering
    \includegraphics[scale=0.45]{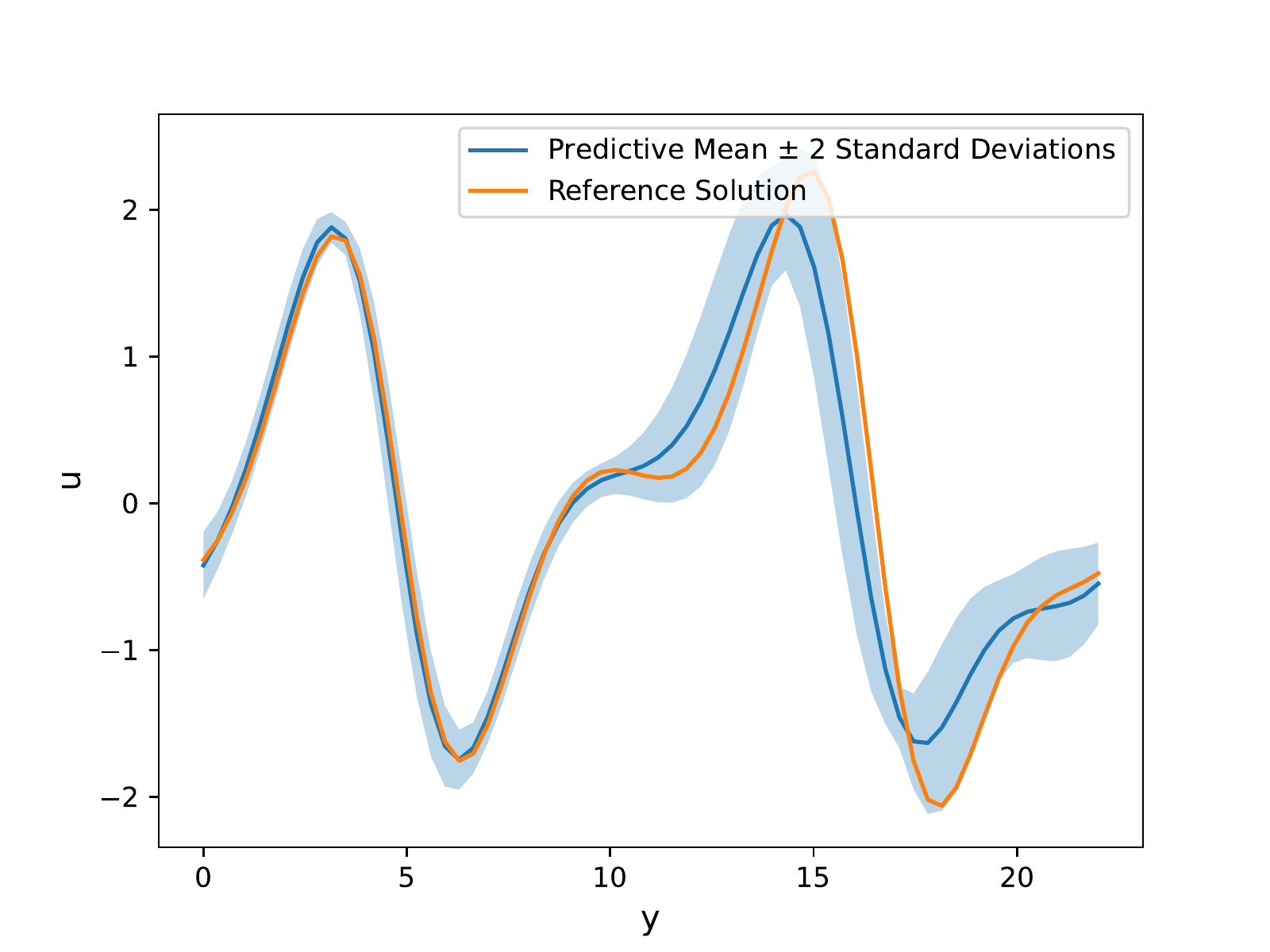}
    \includegraphics[scale=0.45]{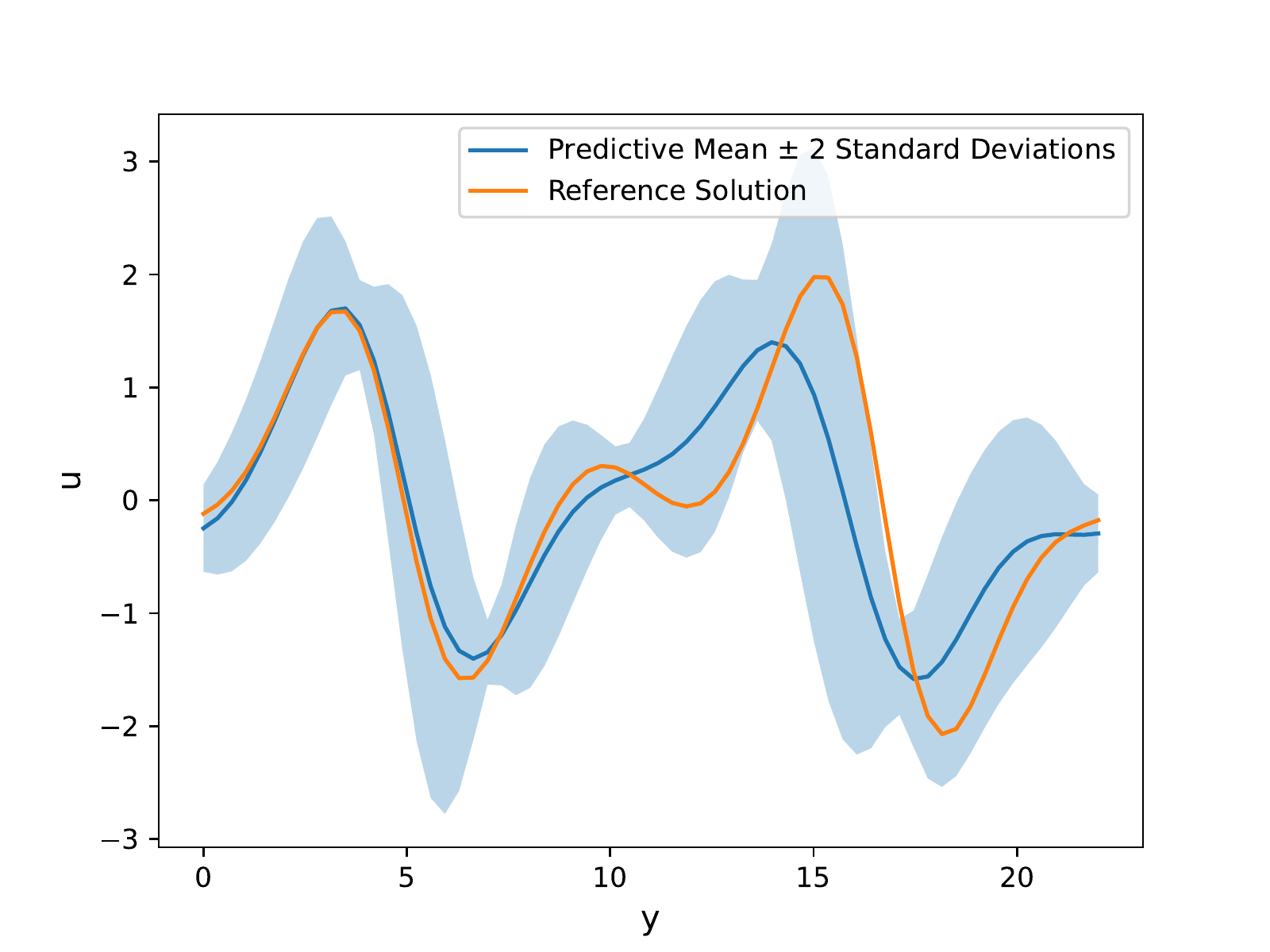}
    \includegraphics[scale=0.45]{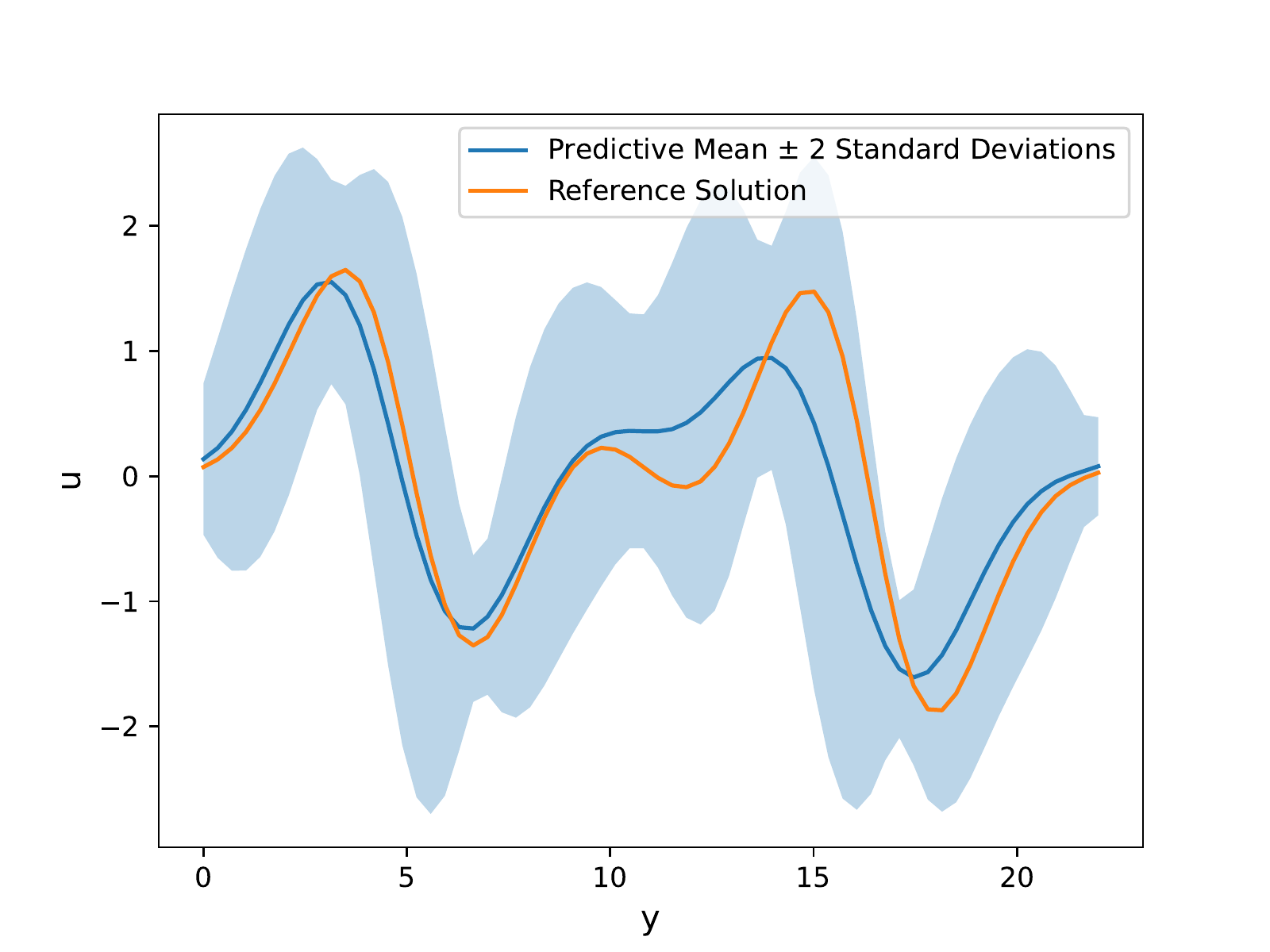}
    \includegraphics[scale=0.45]{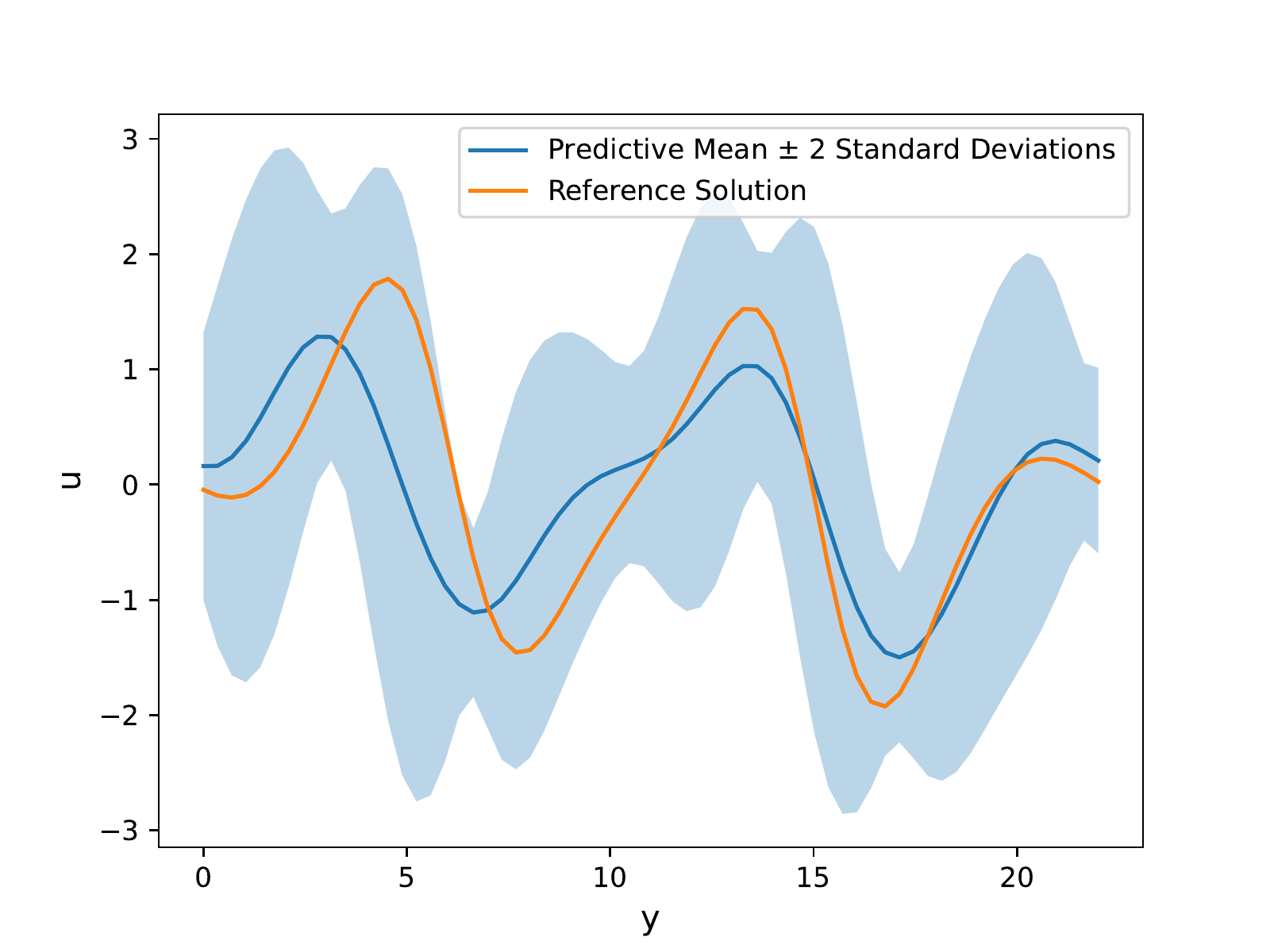}
    \includegraphics[scale=0.45]{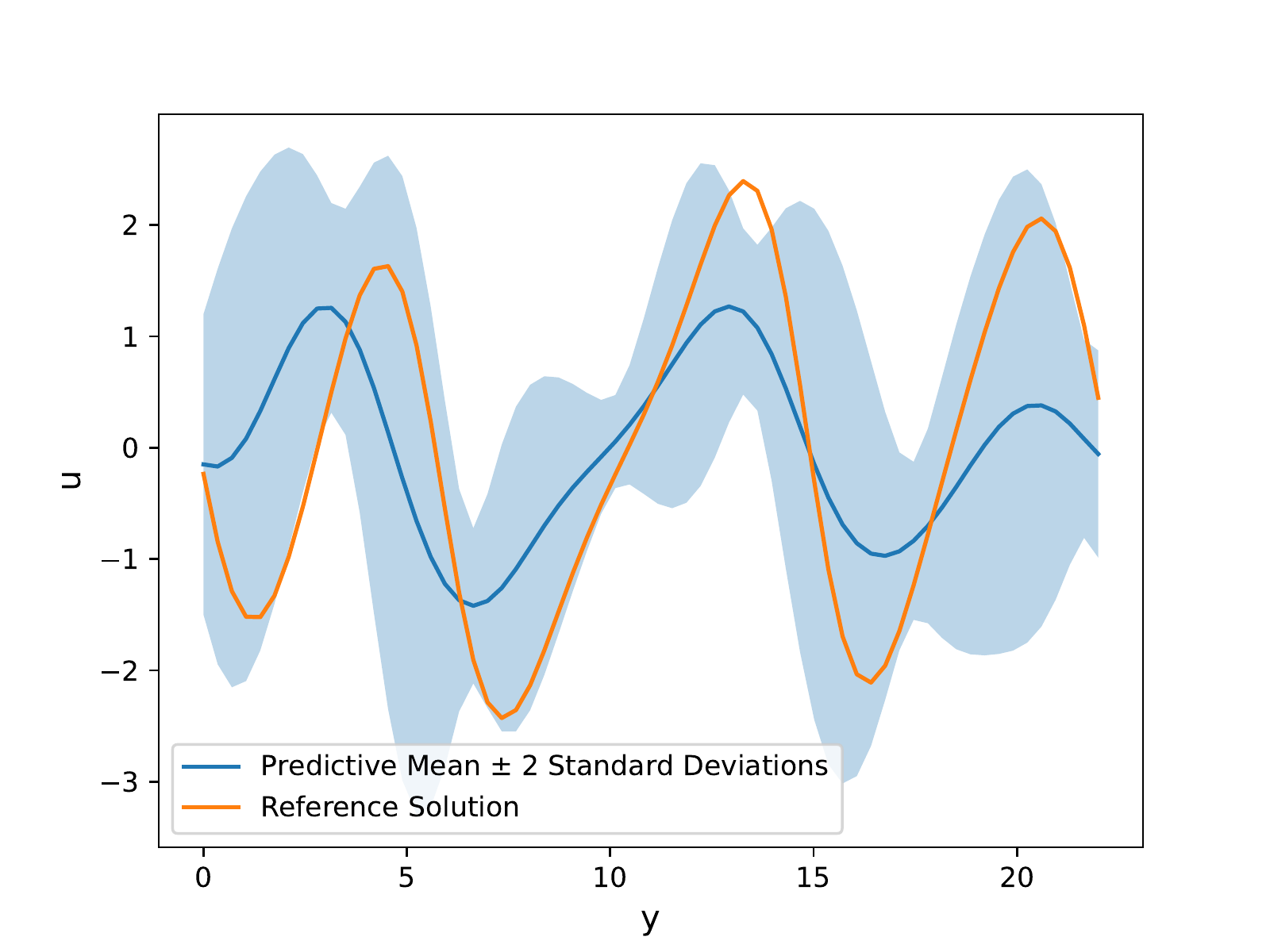}
    \includegraphics[scale=0.45]{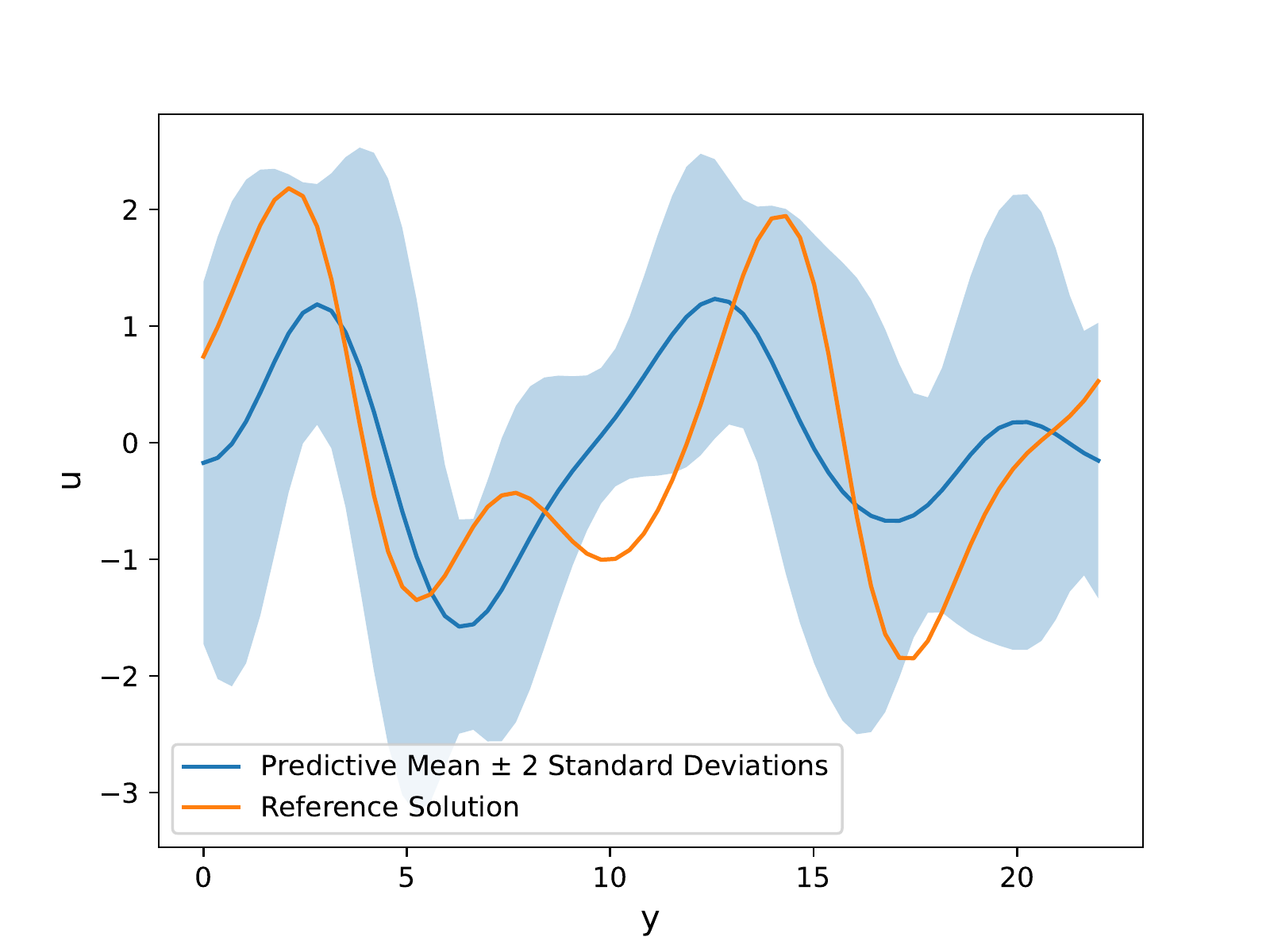}
    \caption{Comparison between predictions and reference solutions for a new initial condition for $t=1.25, 3.75, 7.5, 12.5, 20, 30$ (from left to right and top to down). We note that with longer prediction time the uncertainty bounds increases. Despite the chaotic nature of the KS equation, the predictive posterior mean is close to the reference solution for $t \leq 12.5$  }
    \label{fig:prob}
\end{figure}

The probabilistic model allows us to quantify the uncertainty in predictions. In Figure \ref{fig:prob} predictions for various time-steps and the respective uncertainty bounds are shown for an unseen initial condition. Due to the chaotic nature of the KS-equation and the small amount of training data, the underlying linear dynamic of our model is only able to capture the full dynamics for a limited time horizon. Fortunately, due to the probabilistic approach  the model is capable of capturing chaotic fluctuations with increasingly wide uncertainty bounds.

We also computed the phase space representation for the KS-equation based on the predictions obtained by our model and compare it with the reference solution. The probabilistic model identifies the correct manifold with a better accuracy than the deterministic model. As some of the small-scale fluctuations are accounted as noise, the resulting manifold is more concentrated at the origin and the obtained values are slightly smaller than the reference manifold although their shape is very similar.

\begin{figure}[h]
    \centering
    \includegraphics[scale=0.5]{Figures_KS/Phase_red_KS.pdf}
    \includegraphics[scale=0.5]{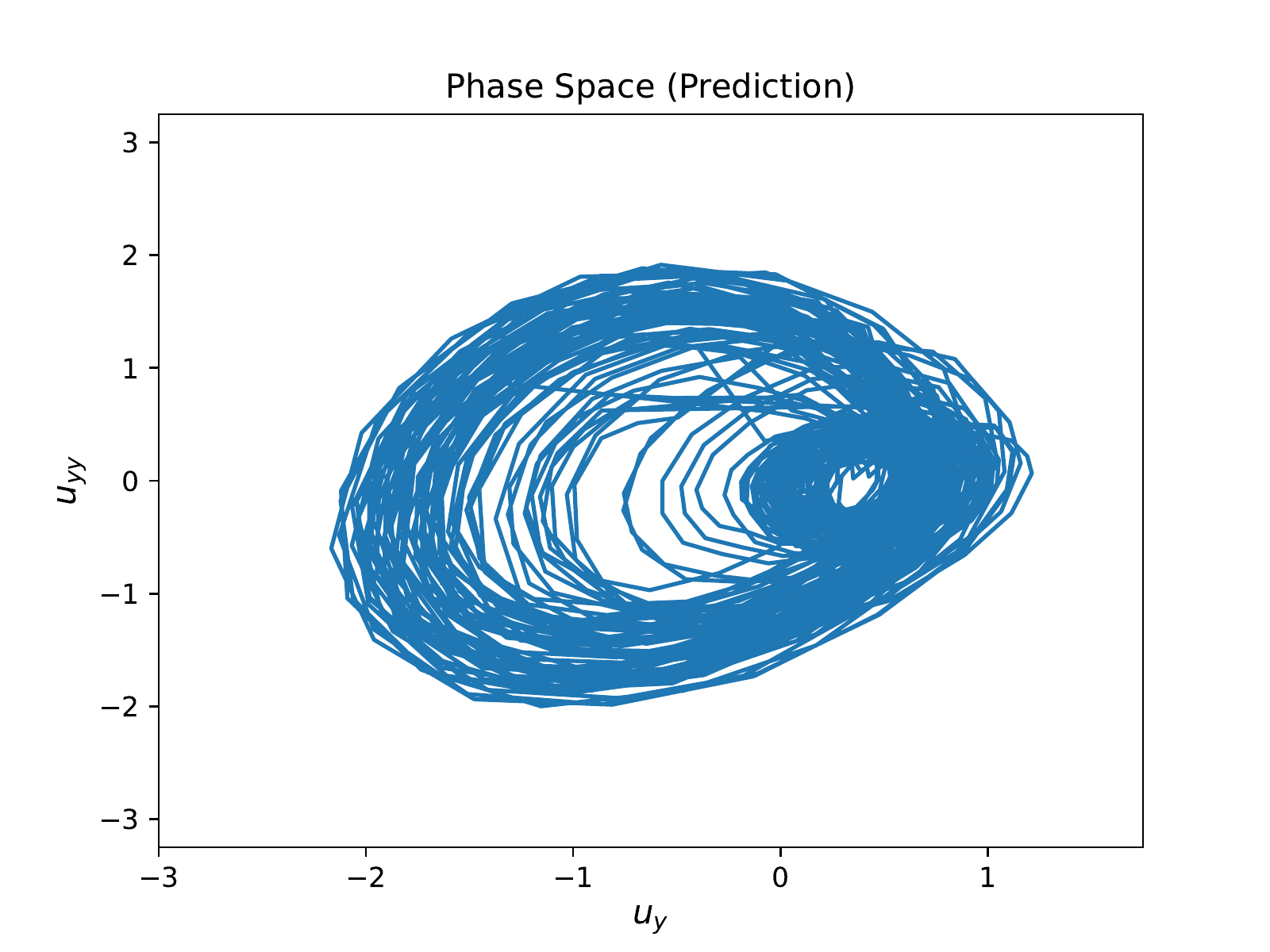}\\
    \includegraphics[scale=0.5]{Figures_KS/Phase_pred_KS.pdf}
    \caption{Comparison between the phase-space of the reference solution (top left), posterior mean of the predictions of the probabilistic model (top right) and phase-space of the predictions of the deterministic model (bottom)}
    \label{fig:my_label}
\end{figure}

\section{Conclusions}
We presented both a deterministic and probabilistic model order reduction framework with interpretable, linear dynamics in the latent space. The parameters associated with the temporal evolution of the latent variables are directly related to the dynamics of the system and indicate increasing/decaying as well as periodic components. The models were applied to three different systems; the probabilistic version was able to additionally identify predictive uncertainties.\\
The interpretable dynamics in are currently linear which is a restriction for our model and thus we are planning on extending our framework such that we are also able to represent non-linear latent dynamics while retaining interpretability.

\section*{Data Availability}
The data that support the findings of this study are available from the corresponding author upon reasonable request.

\section*{Declaration of Interest}
The authors declare no competing interests.

\section*{Acknowledgments}
We acknowledge support by the The European High Performance Computing Joint Undertaking (EuroHPC) Grant DComEX
(956201-H2020-JTI-EuroHPC-2019-1).

\newpage
\appendix
\section{Appendix A}
\label{sec:App_A}
In this appendix we derive, the transition density as well as the necessary condition for a priori stationarity.
For a multivariate Ornstein-Uhlenbeck (OU) process of the form:
\be
d\bz_t=-\bs{A} \bz_t dt+\bs{B} d\bs{W}_t
\label{eq:ou}
\ee
where $\bs{A,B}$ are matrices and $\bs{W}_t$ the multivariate Wiener process, we have that:
\be
\mathbb{E}[\bz_t]=\bs{\mu}_t=e^{-\bs{A}t} \bs{\mu}_0
\label{eq:oum}
\ee
where $\mathbb{E}[\bz_0]=\bs{\mu}_0$.
The stationary covariance matrix $\bs{\Sigma}_{\infty}$ satisfies the Lyapunov equation \citep{gardiner1985handbook}:
\be
\bs{A}\bs{\Sigma}_{\infty}+\bs{\Sigma}_{\infty}\bs{A}^T=\bs{B}\bs{B}^T
\label{eq:ous1}
\ee
and in 2-dimensions is given by \citep{gardiner1985handbook}:
\be
\bs{\Sigma}_{\infty}=\cfrac{det(\bs{A}) \bs{B}\bs{B}^T +(\bs{A}-tr(\bs{A})\bs{I})\bs{B}\bs{B}^T (\bs{A}-tr(\bs{A})\bs{I})^T }{2 tr(\bs{A}) det(\bs{A})}
\label{eq:ous2}
\ee
Lastly the stationary autocorrelation is given by \citep{gardiner1985handbook}:
\be
\mathbb{E}[\bz_{t+\Delta t}\bz_t^T]=e^{-\bs{A} \Delta t} \bs{\Sigma}_{\infty}
\label{eq:oua}
\ee

We consider the complex OU-process $\bz_t=z_{R,t}+iz_{I,t}$ evolving as:
\be
d\bz_t= -\bs{\lambda} \bz_t \; dt + \frac{\sigma}{\sqrt{2}} d\bs{W}_t
\ee
where $\bs{\lambda}=\lambda_R+i\lambda_I$ and  $d\bs{W}_t=dW_{R,t}+idW_{I,t}$ such that $W_{R,t}, W_{I,t}$ are independent Wiener processes. We furter assume that at $t=0$ $\bz_0 \sim \mathcal{CN}(0,\sigma_0^2)$  i.e. $z_{R,0}, z_{I,0}$ are independent, zero-mean normal variables with variance $\sigma^2_0/2$.
If we order the real and imaginary parts of $\bz_t$ in a vector $\bk_t=\left[\begin{array}{c}z_{R,t}\\ z_{I,t}\end{array} \right]$, then the evolution of the 2-dimensional vector is dictated by the OU process of  where:
\bi
\item $\bs{A}=\left[ \begin{array}{cc} \lambda_R & - \lambda_I \\ \lambda_I & \lambda_R \end{array} \right] \longrightarrow tr(\bs{A})=2\lambda_R,~~ det(\bs{A})=|\bs{\lambda}|^2$
\item $\bs{B}=\frac{\sigma}{\sqrt{2}} \bs{I} \longrightarrow \bs{B}\bs{B}^T=\frac{\sigma^2}{2} \bs{I}$
\ei
Based on  we have that:
\be
\bs{\mu}_t=\bs{0}.
\ee
Based on  we have that:
\be
\bs{\Sigma}_{\infty}=\frac{\sigma^2}{4\lambda_R} \bs{I}
\ee
which satisfies .
Based on the  stationary  autocorrelation of  and properties of Gaussian vectors we can derive the stationary transition density $p(\bk_{t+\Delta t} | \bk_t)$ which will also be Gaussian i.e.:
\be
\bk_{t+\Delta t} | \bk_t \sim \mathcal{N}(\bs{\bar{\mu}}_{\Delta t}, \bs{\bar{\Sigma}}_{\Delta t})
\ee
where:
\be
\bs{\bar{\Sigma}}_{\Delta t}=\bs{\Sigma}_{\infty} -e^{-\bs{A} \Delta t} \bs{\Sigma}_{\infty} \bs{\Sigma}_{\infty}^{-1} \bs{\Sigma}_{\infty} e^{-\bs{A}^T \Delta t} = \frac{\sigma^2}{4\lambda_R}(1-e^{-2\lambda_R \Delta t})\bs{I}
\ee
and:
\be
\bs{\bar{\mu}}_{\Delta t}=\bs{0}
+e^{-\bs{A} \Delta t} \bs{\Sigma}_{\infty} \bs{\Sigma}_{\infty}^{-1}(\bs{k}_t -\bs{0})=e^{-\bs{A} \Delta t} \bs{k}_t=e^{-\lambda_R \Delta t}\left[ \begin{array}{cc} \cos(\lambda_I \Delta t) & \sin(\lambda_I \Delta t)\\ & \cos(\lambda_I \Delta t) \end{array} \right] \bk_t
\ee

Hence to obtain a stationary process $\bx_t$ with covariance $\bs{\Sigma}_{\infty}=\frac{1}{2}\bs{I}$ the following parameter choices must be made:
\bi
\item $\sigma^2=2\lambda_R$
\item $\sigma^2_0=1$
\ei

\section{Appendix B}
\label{sec:App_B}
This section contains details of the derivation of the Evidence-Lower-Bound (ELBO) which serves as the objective function for the determination of the parameters $\bp$ and $\bt$ during training. In particular for $N$ given time series $\bx$:

\be
\begin{array}{ll}
 &\log p(\bx_{0:T}^{(1:N)}| \bt)\\
 & =\log  \int p( \bx_{0:T}^{(1:N)},\bz_{0:T}^{(1:N)},\bt ) ~~\bz_{0:T}^{(1:N)}   \\
 & = \log  \int \cfrac{ p( \bx_{0:T}^{(1:N)} | \bz_{0:T}^{(1:N)},\bt) p( \bz_{0:T}^{(1:N)},\bt )}{ q_{\bp}( \bz_{0:T}^{(1:N)})} q_{\bp}( \bz_{0:T}^{(1:N)})  ~d\bz_{0:T}^{(1:N)}  \\
 & \ge \int \log \cfrac{p( \bx_{0:T}^{(1:N)} | \bz_{0:T}^{(1:N)},\bt ) p( \bz_{0:T}^{(1:N)},\bt )}{ q_{\bp}( \bz_{0:T}^{(1:N)})} q_{\bp}( \bz_{0:T}^{(1:N)}) ~d\bz_{0:T}^{(1:N)}  \\
 & = \mathcal{F}(q_{\bp}( \bz_{0:T}^{(1:N)}),\bt)
\end{array}
\label{eq:elboqphi}
\ee

\bibliography{time-scales}

\end{document}